\documentclass[a4paper,fleqn]{cas-sc}

\usepackage[authoryear]{natbib}
\usepackage{amsmath,amssymb}
\usepackage{booktabs}
\usepackage{algorithm}
\usepackage{algpseudocode}
\usepackage{url}
\usepackage{float}

\RenewDocumentCommand \printorcid { } { }
\ExplSyntaxOn
\bool_gset_true:N \g_stm_nologo_bool
\ExplSyntaxOff

\algrenewcommand\algorithmicrequire{\textbf{Input:}}
\algrenewcommand\algorithmicensure{\textbf{Output:}}

\def\tsc#1{\csdef{#1}{\textsc{\lowercase{#1}}\xspace}}
\tsc{UAV}

\begin{document}
\let\WriteBookmarks\relax
\def\floatpagepagefraction{1}
\def\textpagefraction{.001}

\shorttitle{A UAV-Based Multi-Modal Vision System for Automated Sideslope Deformation Monitoring and Hazard Detection}
\shortauthors{Zhang et~al.}

\title [mode = title]{A UAV-Based Multi-Modal Vision System for Automated Sideslope Deformation Monitoring and Hazard Detection}

\author[1]{Jingfeng Zhang}
\ead{zhangjingfeng@m.scnu.edu.cn}

\author[1]{Yi Li}

\author[1]{Xianchong Liang}

\author[1]{Huan Yang}

\affiliation[1]{organization={South China Normal University},
            city={Foshan},
            state={Guangdong},
            country={China}}

\begin{abstract}
Slope hazards constitute a major safety threat to expressway infrastructure, and their evolution is typically manifested as slow surface deformation. Conventional manual inspection suffers from low efficiency and inadequate operational safety, especially on severely deteriorated slopes. Accordingly, there is an urgent need for an automated, high-precision solution capable of large-area slope observation and analysis. This study aims to develop a highly automated workflow for slope hazard detection using Unmanned Aerial Vehicle (UAV)-borne Light Detection and Ranging (LiDAR).  The proposed workflow consists of a shared data-acquisition and ground-surface extraction stage, a single-observation hazard-screening branch based on RandLA-Net, and a multi-epoch deformation-monitoring branch based on grid-wise elevation differencing. To validate the effectiveness of the proposed system, we conducted multiple UAV-borne LiDAR data-acquisition flights in real expressway slope environments. The results show that the workflow can extract usable ground-surface point clouds under vegetation cover, identify potential hazard zones from single-observation point clouds, and quantify centimeter-level elevation changes using multi-epoch grid differencing. This study establishes an end-to-end UAV-borne LiDAR-based workflow for slope inspection and demonstrates its feasibility through controlled experiments, field tests, and simulation-based validation, thereby providing an implementable solution for automated slope-hazard monitoring and intelligent early warning.
\end{abstract}


\maketitle

\section{Introduction}\label{sec:introduction}

As the global expressway network continues to expand, a large proportion of highway infrastructure is undergoing severe aging. As a critical ancillary component of the roadbed, the stability of slopes is directly related to traffic safety and structural durability. Because such slopes remain exposed to natural environmental forcing over long periods, they deteriorate rapidly, and their hazards are often less conspicuous than failures occurring directly on the roadway. Statistics indicate that landslides, collapses, and rockfalls are among the most common slope hazards, and their formation is typically associated with gradual surface deformation. \citet{froude2018global} analyzed a global dataset of fatal non-seismic landslides and reported that 4,862 landslide events caused 55,997 deaths between January 2004 and December 2016; notably, 568 of these events occurred on roads, highlighting the relevance of slope-hazard monitoring to transportation corridors. Under future climate-warming scenarios, moreover, approximately 43.6\%--69.9\% of global road and railway assets are expected to face more frequent extreme rainfall events, substantially increasing slope-instability risk and maintenance costs \citep{liu2023transportation}. Notably, numerous studies have shown that slope failures are typically preceded by progressive early-stage hazards, such as localized settlement, crack propagation, and material loss on the slope surface, all of which are often foreshadowed by subtle ground-elevation changes. If such small deformations can be captured in time, they may provide a critical window for early warning \citep{zhang2023evolution}.

At present, routine condition assessment of expressway slopes still relies primarily on manual on-site visual examination. Under steep or potentially unstable slope conditions, however, such manual fieldwork entails substantial operational risk and cannot efficiently provide large-scale, periodic coverage; early centimeter-level deformation is therefore almost impossible to detect. With the continuous improvement of UAV platform stability and payload capacity, the flexibility and rapid deployability of UAVs have revealed strong potential to replace labor-intensive inspection. Existing studies have widely adopted UAVs equipped with visible-light cameras, using digital photogrammetry to generate high-resolution 3D point clouds and orthophotos for morphological change detection and volumetric analysis, extracting rock-mass discontinuities for rockfall susceptibility assessment, or reconstructing multi-temporal imagery for slope-deformation monitoring. These methods have achieved promising results in bare-rock cuts or sparsely vegetated settings, but in real roadside slope environments, dense vegetation occlusion severely limits the ability of optical imagery to perceive slope defects, constituting an intrinsic limitation that is difficult to overcome.

In contrast, UAV-borne LiDAR can acquire ground returns beneath vegetation by emitting laser pulses and recording multiple returns from canopy and ground surfaces. With multi-return capability and multi-angle scanning, such systems can improve ground-point coverage in vegetated slope environments and mitigate the limitations of optical photogrammetry. For change detection, multi-temporal point-cloud differencing methods, such as the M3C2 algorithm, can detect topographic change \citep{lague2013m3c2} at the millimeter-to-centimeter level without requiring DEM gridding. For hazard recognition, PointNet pioneered a direct deep-learning approach \citep{qi2017pointnet} for processing unordered 3D point clouds. RandLA-Net further improved the scalability \citep{hu2020randla} of point-cloud semantic segmentation by combining random sampling with local feature aggregation. This enables efficient processing of large-scale outdoor point clouds and provides a practical technical basis for automated semantic recognition in expressway slope scenes.

Nevertheless, despite the demonstrated advantages of UAV-borne LiDAR in individual tasks, engineering practice still lacks an end-to-end automated workflow that integrates standardized data acquisition, ground-surface extraction, single-observation hazard screening, and multi-epoch deformation monitoring. Most previous studies have focused on only one of these components and therefore cannot support a truly end-to-end monitoring workflow from raw data to actionable hazard insight. In addition, although efficient point-cloud segmentation models such as RandLA-Net have performed well in generic outdoor scenes, their systematic application and engineering validation in real expressway slope-hazard detection remain largely unexplored.

To address these gaps, this study proposes and develops an end-to-end UAV-borne LiDAR-based inspection workflow for expressway slopes. Its main contributions are as follows:

\begin{enumerate}
\item We establish an end-to-end UAV-borne LiDAR-based workflow that converts raw UAV-borne LiDAR data into interpretable slope-hazard outputs through standardized data acquisition, ground-surface extraction, single-observation hazard screening, and multi-epoch deformation monitoring.
\item We develop two complementary analytical branches based on the shared ground-surface representation: a RandLA-Net-based single-observation hazard-screening branch for no-baseline inspection scenarios, and a multi-epoch grid-differencing branch for quantifying progressive elevation changes between time-separated UAV-borne LiDAR observations.
\item We validate the proposed workflow through an under-canopy ground-surface extraction experiment, simulation-based single-observation hazard screening, controlled known-deformation testing, and real expressway slope field tests, demonstrating its practical feasibility for automated slope inspection and hazard-related output generation.
\end{enumerate}

\begin{figure}[pos=H]
  \centering
  \includegraphics[width=\figwidth,height=.68\textheight,keepaspectratio]{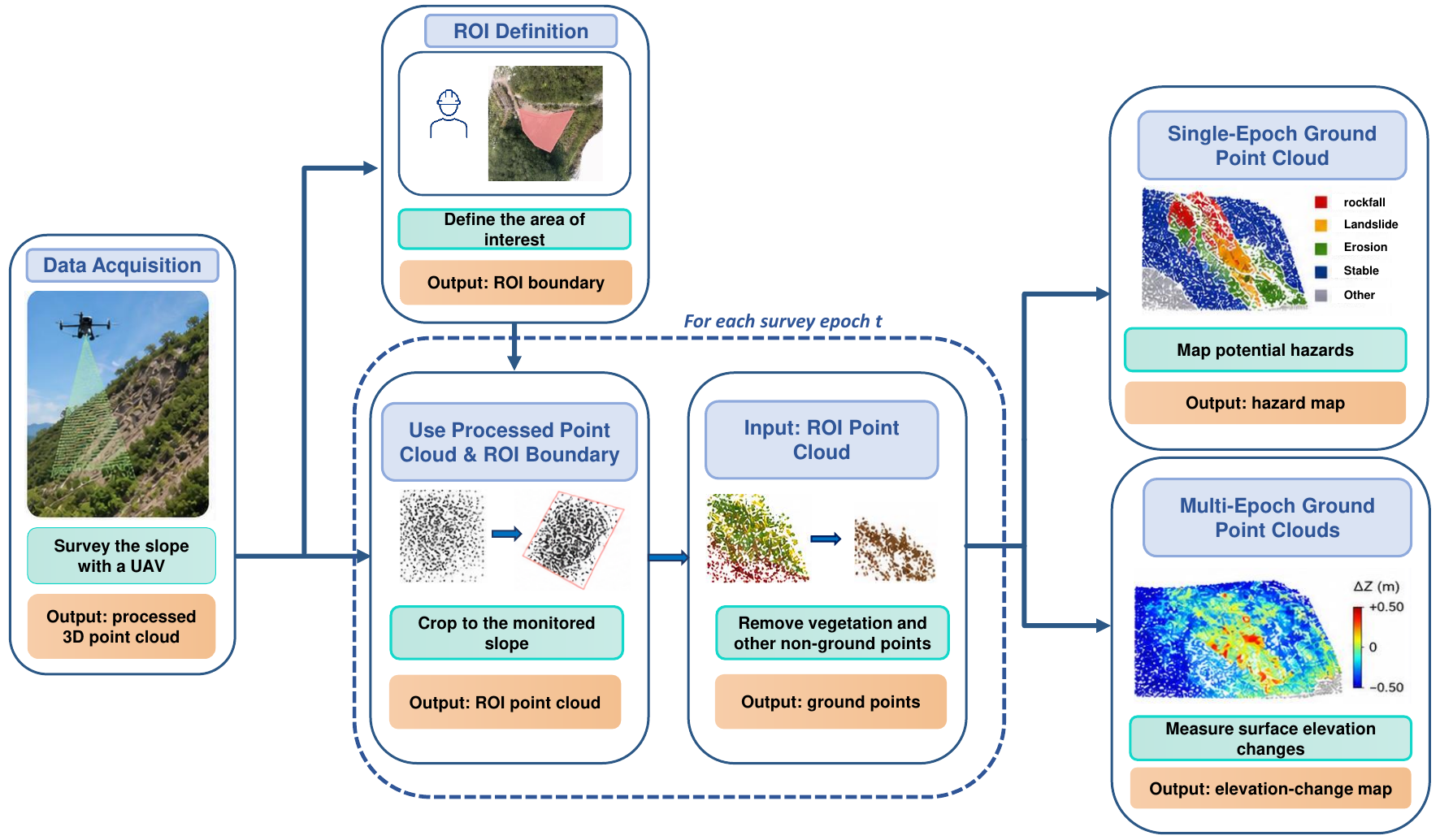}
  \caption{Overall workflow of the proposed UAV-borne LiDAR-based slope inspection system.}\label{fig:1}
\end{figure}

\section{Related Work}\label{sec:related-work}

\subsection{Common Slope Hazards and Their Monitoring Requirements}

Highway slopes are exposed to long-term rainfall erosion, geological weathering, and engineering disturbance, making them susceptible to various hazards. Existing studies generally classify slope hazards into landslides, collapses or rockfalls, erosion-induced damage, surface subsidence, and progressive slope deformation \citep{chen2025roadbed}. Among them, landslides and collapses may directly cause traffic interruption and casualties, whereas erosion, weathering, shallow instability, and subsidence often weaken the slope mass progressively and accelerate the development of larger-scale instability.

Large-scale slope failure is usually not an isolated sudden event, but the final stage of a progressive deformation process. Early-stage hazards may appear as toe deformation, retaining-structure displacement, tensile cracks, surface settlement, erosion depressions, or local material loss, and are often associated with centimeter- to decimeter-scale surface-elevation changes \citep{zhang2023evolution,wu2025rainwater}. Cracks also provide important evidence for identifying potential landslide boundaries, although LiDAR alone may still be limited for continuous time-series interpretation in some scenarios \citep{deng2025cracks}. Therefore, effective slope monitoring requires high-resolution three-dimensional data that can capture subtle geometric, roughness, and structural changes over large, steep, and often vegetated areas. This requirement motivates the use of UAV-borne LiDAR, which has stronger ground-perception capability under vegetation cover than conventional manual inspection or optical imagery.

\subsection{UAV-Based Slope Inspection Methods}

\begin{figure}[pos=H]
  \centering
  \includegraphics[width=\figwidth,height=.68\textheight,keepaspectratio]{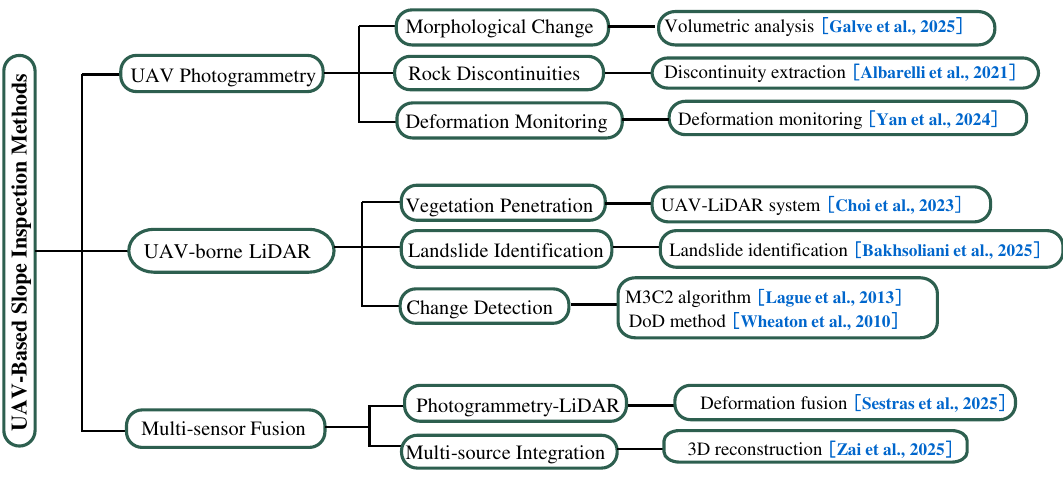}
  \caption{Literature organization of UAV-based slope inspection methods.}\label{fig:2}
\end{figure}

With the development of UAV platforms and onboard sensors, UAV-based inspection has become an important technical route for slope-hazard investigation. As summarized in Fig.~\ref{fig:2}, existing studies mainly follow two sensing directions: UAV photogrammetry and UAV-borne LiDAR, with further extensions toward multi-epoch change detection and multisensor fusion. UAV photogrammetry reconstructs high-resolution orthophotos and 3D point clouds from visible-light imagery, and has been widely applied to slope morphological change analysis, volume estimation, rock-mass discontinuity extraction, and multitemporal deformation monitoring \citep{sun2024uavs,galve2025uav,albarelli2021rockfall,yan2024uav}. However, optical imagery mainly captures canopy and exposed surface radiance. In densely vegetated expressway slopes, the true ground surface may be obscured, making photogrammetric methods less effective for detecting subsidence, sliding, or erosion beneath vegetation.

Compared with photogrammetry, UAV-borne LiDAR actively emits laser pulses and records multiple returns, allowing partial penetration through vegetation canopy and acquisition of ground-surface points. \citet{choi2023uavlidar} demonstrated that UAV LiDAR can obtain ground point clouds in vegetated areas, while other studies have used LiDAR-derived DEMs and terrain factors such as slope, curvature, terrain roughness index (TRI), and topographic position index (TPI) to identify landslide boundaries and morphological features under mixed vegetation cover \citep{bakhsoliani2025surami}. These studies show that UAV-borne LiDAR is particularly suitable for expressway slope environments where protective vegetation often limits optical ground observation.

For deformation monitoring, multi-epoch point-cloud comparison has become a key approach for quantifying slope evolution. DEM differencing (DoD) and direct point-cloud comparison methods such as M3C2 have been widely used for topographic-change analysis, with M3C2 reducing some geometric errors introduced by DEM gridding \citep{wheaton2010dod,lague2013m3c2,chen2025uavreview}. Multisensor fusion further combines the complementary advantages of photogrammetry, terrestrial laser scanning, and airborne LiDAR for deformation monitoring in complex terrain \citep{sestras2025fusion,zai2025multisource}. Overall, existing studies have made progress in UAV data acquisition, LiDAR-based terrain perception, point-cloud differencing, and multisensor fusion. However, most of them remain focused on individual components and have not fully integrated standardized acquisition, vegetation filtering, ground-surface extraction, multi-epoch deformation analysis, and deep-learning-based hazard recognition into a unified automated workflow for expressway slope inspection.

\subsection{Advances in Deep Learning for Three-Dimensional Point Clouds}

Deep learning has provided new opportunities for automated geological-hazard recognition. Previous studies have applied deep learning to landslides, debris flows, collapses, and related hazards using remote-sensing images, UAV data, and other multisource observations \citep{ma2021deeplearning}. Compared with two-dimensional images, three-dimensional point clouds directly describe surface geometry and are therefore more suitable for representing microtopographic changes, cracks, local uplift, subsidence, and deformation-related structures. Recent reviews also indicate that high-resolution topographic data, including LiDAR-derived point clouds, are increasingly used in deep-learning-based landslide analysis \citep{jiang2026review}.

\begin{figure}[pos=H]
  \centering
  \includegraphics[width=\figwidth,height=.68\textheight,keepaspectratio]{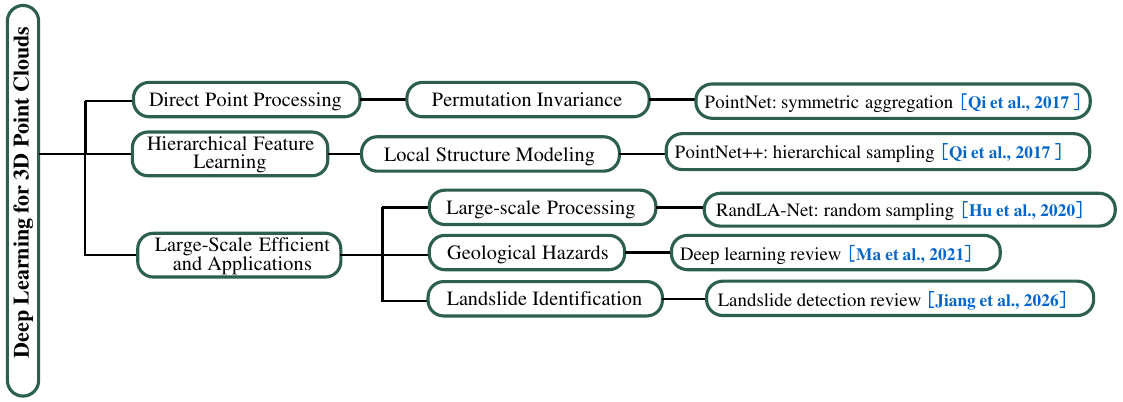}
  \caption{Literature organization of deep learning methods for three-dimensional point clouds.}\label{fig:3}
\end{figure}

As summarized in Fig.~\ref{fig:3}, point-cloud deep learning has evolved from direct point-set processing to more efficient large-scale semantic segmentation. The main challenge is that LiDAR point clouds are unordered and irregular, lacking the regular grid structure required by conventional convolutional neural networks. PointNet first enabled direct learning from unordered point sets through symmetric aggregation, but its ability to model local geometric structures was limited \citep{qi2017pointnet}. PointNet++ improved local feature learning through hierarchical sampling and neighborhood aggregation, but its computational cost remains high for large outdoor scenes \citep{qi2017pointnetplusplus}. RandLA-Net further improved scalability by combining random sampling with local feature aggregation, enabling efficient semantic segmentation of large-scale outdoor point clouds \citep{hu2020randla}. Nevertheless, these models have mainly been validated in general outdoor scenes such as urban roads and forested environments. Their systematic application to highway slope hazards remains limited because slope scenes involve persistent inclination, strong background geometric trends, and subtle local deformation signals. This gap motivates the use and adaptation of efficient point-cloud semantic segmentation methods for UAV-borne LiDAR-based slope-hazard screening.

\section{Problem Statement}\label{sec:problem}

\subsection{Engineering Task and Two Observation Scenarios}

Expressway slope hazards usually evolve through progressive surface deformation rather than instantaneous failure, making timely geometric observation important for early warning and maintenance prioritization. Therefore, this study formulates the central problem as how an end-to-end UAV-borne LiDAR workflow can transform raw point clouds into reliable and interpretable information for slope-hazard judgment. In practical expressway maintenance, this task does not occur under a single observation condition. Instead, it involves two typical engineering scenarios with different data availability and analytical requirements.

Let the raw UAV-borne LiDAR point cloud acquired at epoch $t$ be denoted as $P_{\mathrm{raw}}^{(t)}$. The overall problem is to construct a workflow $\mathcal{W}$ that maps raw observations to hazard-related outputs:

\[
\mathcal{W}: P_{\mathrm{raw}}^{(t)} \ \mathrm{or}\ \{P_{\mathrm{raw}}^{(1)},P_{\mathrm{raw}}^{(2)}\} \rightarrow \{Y^{(t)}, H_{\Delta Z}\},
\]

where $Y^{(t)}$ denotes the single-observation semantic hazard map and $H_{\Delta Z}$ denotes the multi-epoch elevation-difference heatmap.

The first scenario is single-observation hazard screening, where hazard judgment must rely on a single-epoch point cloud because historical LiDAR observations are unavailable, inconsistent, or unsuitable for direct comparison. The task is to identify slope regions that already exhibit morphological anomalies, such as local uplift, subsidence, or erosion-induced depressions, using the spatial information contained in the current UAV-borne LiDAR observation.

The second scenario is multi-epoch deformation monitoring based on time-separated UAV-borne LiDAR observations. Once repeated observations are available, a later epoch can be compared with an earlier baseline epoch to detect slow and subtle surface-elevation changes on the same slope, thereby localizing areas of progressive deformation that may not yet be visually obvious.

These two scenarios are complementary: single-observation screening supports first-inspection or no-baseline cases, whereas multi-epoch analysis supports quantitative deformation tracking once repeated observations are available. Both scenarios, however, depend on a reliable ground-surface representation extracted from UAV-borne LiDAR point clouds.

\begin{figure}[pos=H]
  \centering
  \includegraphics[width=\figwidth,height=.68\textheight,keepaspectratio]{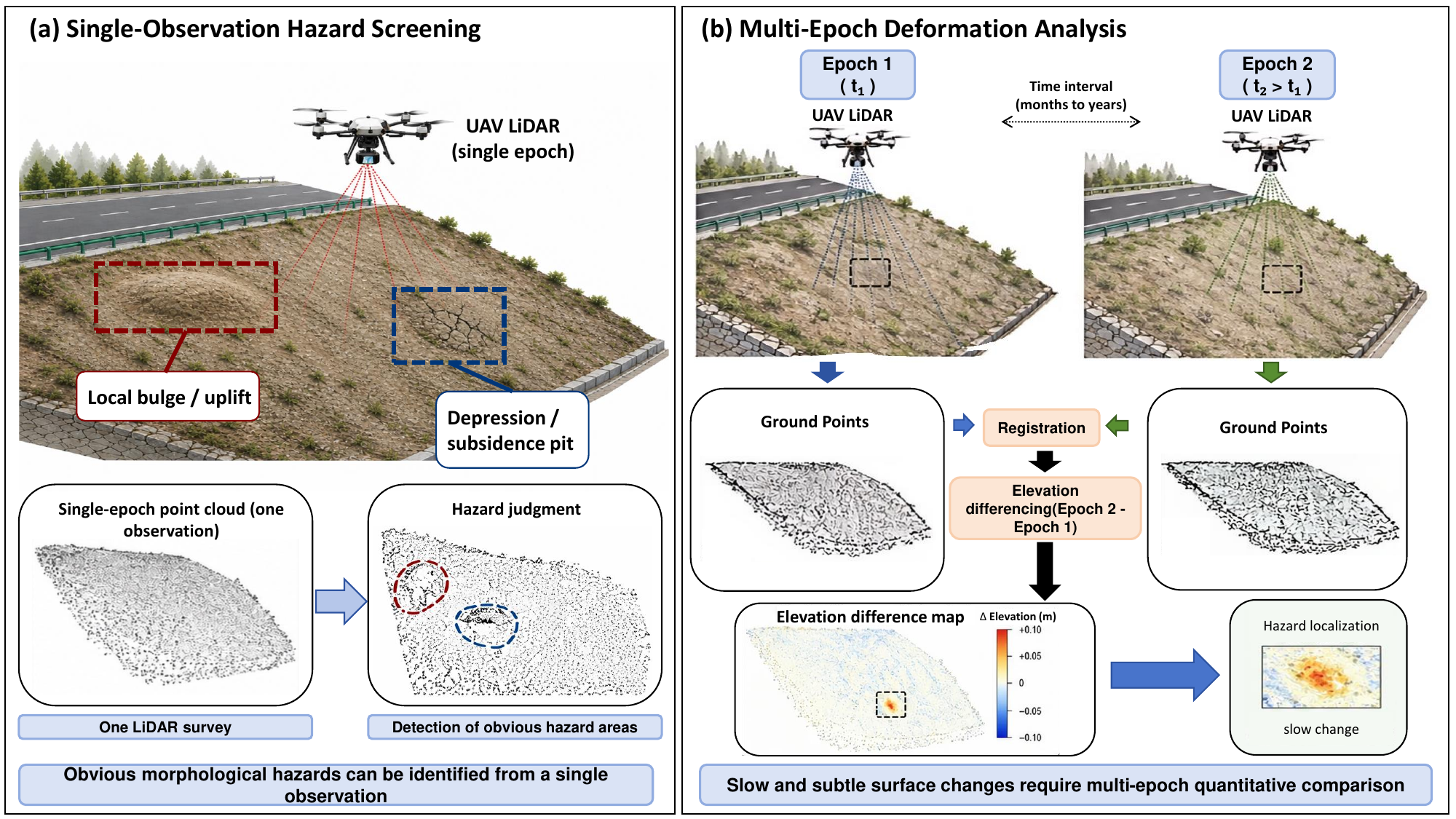}
  \caption{Problem formulation of the two observation scenarios, showing single-observation hazard screening and multi-epoch deformation monitoring for expressway slope inspection.}\label{fig:4}
\end{figure}

Fig.~\ref{fig:4} illustrates these two observation scenarios, including obvious morphological hazards that can be judged from a single observation and slow surface changes that require multi-epoch observations for quantitative comparison.

\subsection{Ground-Surface Representation under Vegetation Cover}

Both observation scenarios require a reliable representation of the slope ground surface, but this requirement is difficult to satisfy in expressway environments because slopes are frequently covered by turf, shrubs, and trees. Such vegetation may stabilize the slope surface, yet it also blocks optical observation and mixes canopy, branch, and ground returns in the acquired point cloud. As a result, deformation occurring beneath vegetation cannot be interpreted reliably from the raw mixed point cloud alone.

UAV-borne LiDAR is more suitable for this task than visible-light photogrammetry because it actively emits laser pulses and records multiple returns from vegetation and ground surfaces. Nevertheless, dense vegetation still reduces the number of valid ground returns and introduces non-ground contamination. Therefore, the problem requires extraction of a usable ground-surface point cloud from raw UAV-borne LiDAR data before either single-observation hazard screening or multi-epoch deformation monitoring can be performed.

This requirement can be expressed as separating the raw mixed point cloud into ground and non-ground subsets,

\[
P_{\mathrm{raw}} = P_{\mathrm{g}} \cup P_{\mathrm{ng}},
\]

where $P_{\mathrm{g}}$ is the ground-surface subset required by downstream analysis and $P_{\mathrm{ng}}$ contains vegetation and other above-ground returns.

\subsection{Single-Observation Hazard Screening without Historical Baseline}

When no historical baseline is available, single-observation hazard screening must infer potential hazard areas from the spatial geometry of the current UAV-borne LiDAR point cloud. The objective is to identify slope regions that already exhibit abnormal morphological characteristics, such as local uplift, subsidence, pits, or erosion-induced depressions, and to produce a spatially interpretable hazard map from one observation.

For a ground-surface point cloud $P_{\mathrm{g}}^{(t)}=\{p_i^{(t)}\}_{i=1}^{N}$, the task is to estimate point-wise semantic labels

\[
f_{\theta}:P_{\mathrm{g}}^{(t)} \rightarrow Y^{(t)}=\{y_i^{(t)}\}_{i=1}^{N},
\]

where $y_i^{(t)}$ represents the class of point $p_i^{(t)}$, such as normal background, uplift hazard, or subsidence hazard. This scenario therefore requires point-level recognition of local geometric anomalies from a single ground-surface point cloud, while avoiding confusion between true hazard-related morphology and the normal inclined background of the slope.

\subsection{Multi-Epoch Deformation Monitoring}

When time-separated UAV-borne LiDAR observations are available, the objective is to quantify temporal surface-elevation changes that may indicate progressive slope deformation. The required output is a spatially interpretable elevation-difference map that identifies where the slope surface has risen, subsided, or remained stable between epochs.

This scenario cannot rely on direct point-to-point subtraction because independently acquired LiDAR point clouds may have small coordinate-frame offsets and do not sample exactly the same ground locations. Therefore, the problem requires consistent inter-epoch alignment and comparison over spatially corresponding surface units.

After two epochs are aligned and converted to common grid cells, the monitoring target can be expressed as

\[
\Delta Z_{i,j}=\bar{Z}_{i,j}^{(2)}-\bar{Z}_{i,j}^{(1)}, \quad (i,j)\in K^{(1)}\cap K^{(2)},
\]

where $\bar{Z}_{i,j}^{(1)}$ and $\bar{Z}_{i,j}^{(2)}$ are mean ground elevations in the same valid grid cell at two epochs. This scenario requires a comparable spatial representation that can separate true surface-elevation change from acquisition-induced inconsistency between epochs.

\section{Proposed Method}\label{sec:method}

\subsection{Overall Workflow Design}

\begin{figure}[pos=H]
  \centering
  \includegraphics[width=\figwidth,height=.68\textheight,keepaspectratio]{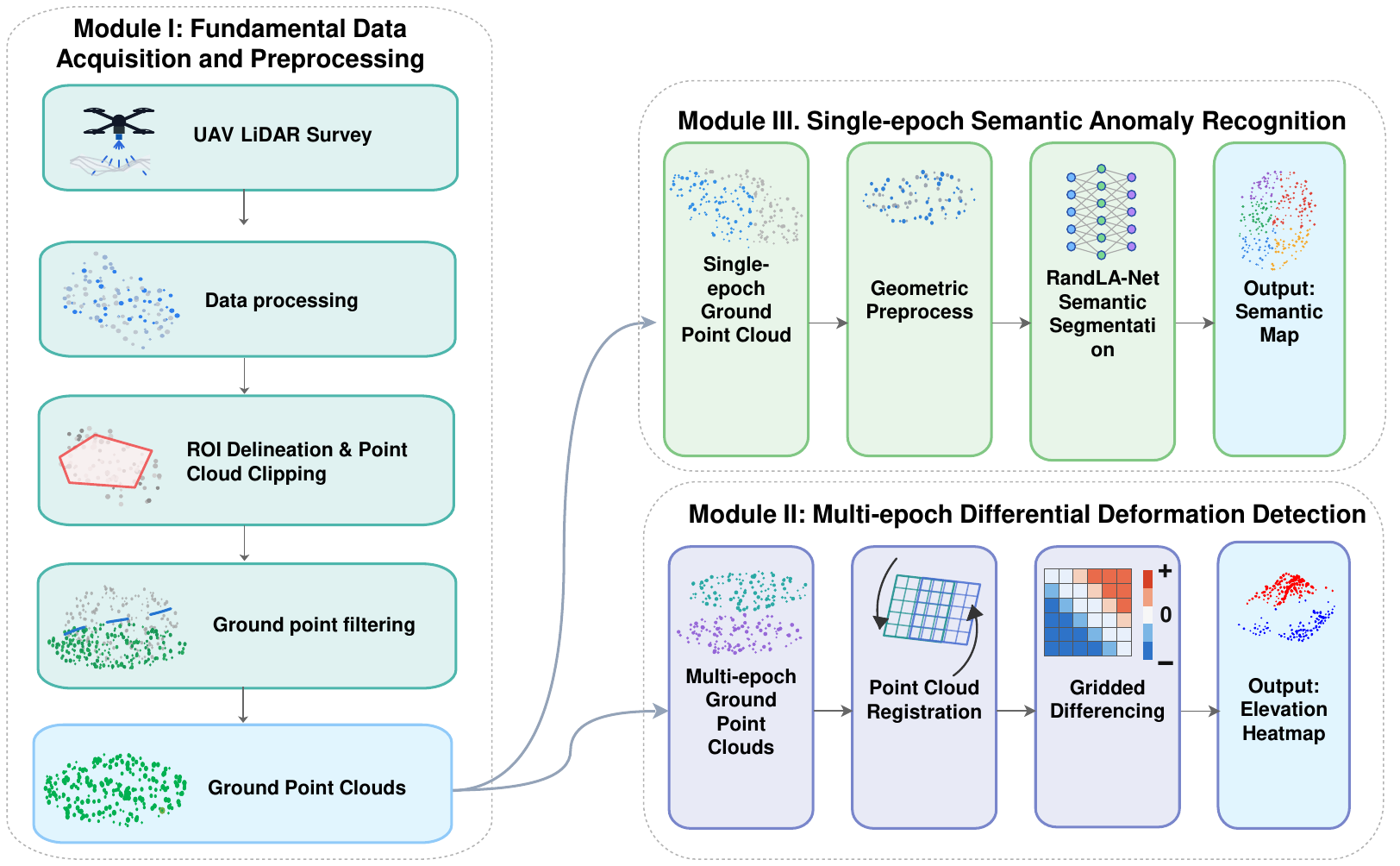}
  \caption{Overview of the proposed method, consisting of shared UAV LiDAR data acquisition and preprocessing, single-epoch semantic anomaly recognition, and multi-epoch differential deformation detection.}\label{fig:5}
\end{figure}

To implement the problem structure defined in Section~\ref{sec:problem}, this study proposes an end-to-end UAV-borne LiDAR-based workflow for expressway slope inspection, as shown in Fig.~\ref{fig:5}. The workflow transforms raw UAV-borne LiDAR data into hazard-related outputs through one shared preprocessing stage and two analysis branches.

The shared preprocessing stage processes the raw point clouds through UAV-borne LiDAR acquisition, trajectory solution, point-cloud reconstruction, region-of-interest delineation, point-cloud clipping, and ground-surface extraction. Its output is a slope ground-surface point cloud, which serves as the input for subsequent analysis.

Based on this ground-surface representation, the workflow contains two branches. The first branch performs single-observation hazard screening using a RandLA-Net-based point-wise semantic recognition module. The second branch performs multi-epoch deformation monitoring through point-cloud registration, grid construction, common-grid intersection, and elevation differencing. The outputs of the two branches are, respectively, a point-wise semantic segmentation map and an elevation-difference heatmap.

The following subsections introduce the shared data acquisition and preprocessing stage, the single-observation hazard screening branch, and the multi-epoch deformation monitoring branch.

\subsection{UAV-borne LiDAR Acquisition and Ground-Surface Extraction}

The shared preprocessing stage is designed to convert raw UAV-borne LiDAR observations into slope ground-surface point clouds for subsequent analysis. It consists of four main steps: UAV-borne LiDAR survey, point-cloud reconstruction, region-of-interest delineation and clipping, and ground-surface extraction. The output of this stage is used as the common input for both the single-observation hazard-screening branch and the multi-epoch deformation-monitoring branch.

Following the problem definition in Section~\ref{sec:problem}, this stage estimates a ground-surface subset from a mixed point cloud:

\[
\mathcal{G}:P_{\mathrm{raw}} \rightarrow P_{\mathrm{g}}.
\]

In vegetated slopes, the non-ground subset $P_{\mathrm{ng}}$ may dominate the raw observation, while the valid ground returns are sparse and unevenly distributed. Therefore, this stage is not only a data-format conversion step, but also the procedure that determines whether the subsequent semantic recognition and temporal differencing branches operate on a physically meaningful slope surface.

For a monitored slope area $A$, vegetation coverage can be expressed as

\[
C_v(A)=\frac{\mathrm{Area}_{\mathrm{veg}}(A)}{\mathrm{Area}(A)} \in [0,1].
\]

As $C_v(A)$ increases, the visible ground surface becomes progressively obscured and the proportion of non-ground returns increases. This explains why the workflow first separates ground and non-ground returns before performing either hazard recognition or elevation differencing.

\subsubsection{UAV-borne LiDAR Survey and Flight-Path Planning}

For each specific slope, we first delineate the area to be managed and then design a corresponding flight path to ensure complete coverage of the target inspection area. To guarantee inter-epoch consistency, the same flight path is maintained for repeated surveys over the same slope to improve spatial consistency between epochs.

The LiDAR scan pattern is selected according to the need for ground-return acquisition under vegetation cover. This study adopts a non-repetitive scanning mode, which provides a wider vertical field of view and multi-angle beam emission within a single flight, thereby improving ground-point coverage in occluded slope regions.

Detailed descriptions of UAV-borne LiDAR parameter settings, including scanning mode, return number, sampling frequency, and point-cloud recording procedures, are provided in the Zenmuse L2 Operation Guide \citep{dji2024zenmusel2}.

\subsubsection{Point-Cloud Reconstruction and LAS Export}

After data acquisition, the UAV-borne LiDAR data are processed through trajectory solution and point-cloud reconstruction, and the resulting point clouds are exported in LAS format \citep{asprs2025las}. Each point contains three-dimensional coordinates $(X,Y,Z)$, intensity, return-number information, and RGB attributes. These attributes provide the basic data support for subsequent ground-point extraction, visualization, and deformation analysis.

\subsubsection{Region-of-Interest Delineation and Point-Cloud Clipping}

Because raw UAV-acquired point clouds usually include roads, guardrails, green belts, pavements, and other irrelevant objects, the region of interest (ROI) must be delineated before slope analysis. The raw 3D point cloud is projected onto the horizontal plane to generate a top-view image, which presents the planar distribution of slopes, roads, guardrails, and surrounding objects. This projection provides the basis for visual ROI annotation.

In addition, point density near the outer boundary of the acquisition range is usually lower than that in the central region because of flight-path and scan-angle constraints. These peripheral edge areas are also more affected by noise and data loss. Therefore, sparse and unreliable edge zones are excluded during ROI delineation, and only regions with relatively uniform point density and reliable data quality are retained.

Professional annotators then use platforms such as Label Studio to annotate the top-view projection and delineate the managed slope region segment by segment according to administrative documents and field investigation results. The annotation generates a polygonal ROI mask representing the target jurisdictional area.

After obtaining the polygonal ROI boundary, the original point cloud is clipped by checking whether each point falls inside the ROI in the horizontal plane. Points outside the managed slope region are removed, while the full 3D elevation information of the retained points is preserved. This step reduces irrelevant data and computational load for subsequent processing.

\subsubsection{Ground-Surface Extraction from Mixed Returns}

To reduce the interference of vegetation and other non-ground objects, the Cloth Simulation Filter (CSF) is applied \citep{zhang2016csf} to the clipped point cloud for ground-point extraction. The basic idea of CSF is to invert the point cloud and simulate an elastic cloth falling onto the terrain surface. Under simulated gravity and cloth-rigidity constraints, the cloth gradually conforms to the overall terrain shape.

After the simulated cloth surface is generated, each original point is classified according to its vertical distance from the cloth surface. Points with smaller distances are labeled as ground points, whereas points with larger distances are identified as non-ground points, such as vegetation, buildings, or other elevated objects.

Several parameters control the CSF result, including cloth resolution, cloth rigidity, time step, iteration number, and classification threshold. Cloth resolution determines the fineness of the simulated cloth; cloth rigidity controls its adaptability to complex terrain; the time step and iteration number affect convergence; and the classification threshold determines the maximum point-to-cloth distance allowed for ground-point labeling.

The extracted ground points form the slope ground-surface point cloud used by the two downstream branches. This output provides the geometric basis for single-observation semantic segmentation and the comparable surface representation required for multi-epoch elevation differencing.

\subsection{Single-Observation Hazard Screening Branch}

\subsubsection{Functional Role and Input--Output Definition}

The single-observation branch takes one slope ground-surface point cloud as an independent input and performs point-wise semantic recognition of potential hazard regions. This branch is used when historical baseline data are unavailable or unsuitable for temporal comparison. Its input is the ground-surface point cloud extracted from the shared preprocessing stage, where each point uses only its three-dimensional coordinate $(x,y,z)$ as the feature. The output is a semantic label for each ground point, including ``normal background,'' ``uplift hazard,'' and ``subsidence hazard.'' The predicted labels are visualized as a point-wise semantic segmentation map to support localization of potential hazard regions.

\subsubsection{Model Selection: RandLA-Net}

As discussed in the related work, different point-cloud deep learning models have different strengths in local geometric representation, computational efficiency, and large-scale scene processing. In this study, RandLA-Net is adopted as the backbone network for the single-observation hazard screening branch because it provides a practical balance between efficiency and point-wise semantic segmentation capability for large-scale outdoor point clouds.

RandLA-Net replaces computationally expensive sampling strategies with random sampling, which substantially reduces the computational burden when processing large point clouds. To compensate for the possible loss of local geometric information caused by random sampling, it introduces a local spatial encoding module (LocSE) to explicitly encode relative point positions and neighborhood geometry. The attentive pooling mechanism further assigns adaptive weights to neighboring features, allowing the network to emphasize more informative local structures during feature aggregation.

In addition, RandLA-Net adopts an encoder--decoder architecture with skip connections. The encoder progressively enlarges the receptive field through repeated sampling and local feature aggregation, while the decoder restores point-level resolution for semantic prediction. This structure enables efficient processing of large-scale point clouds while maintaining point-wise output, which matches the requirements of coordinate-only slope hazard recognition in this study.

\subsubsection{Data Preprocessing Pipeline}

The preprocessing pipeline is motivated by the geometric coupling between the inclined slope background and local hazard deformation. A local anomaly can be regarded as a small perturbation superimposed on the dominant slope trend:

\[
z(x,y)=z_{\mathrm{slope}}(x,y)+\delta(x,y),
\]

where $z_{\mathrm{slope}}(x,y)$ represents the macroscopic inclined surface and $\delta(x,y)$ represents local uplift, subsidence, or erosion. The relative strength of the local anomaly can be approximated by a signal-to-background ratio:

\[
\mathrm{SBR}=\frac{\|\delta\|}{|\tan \alpha|\cdot \|\Delta x\|},
\]

where $\alpha$ is the slope angle and $\Delta x$ is the monitored downslope extent. When the background elevation variation dominates the local perturbation, $\mathrm{SBR}$ becomes small and absolute coordinates alone may obscure the hazard signal. Therefore, the single-observation branch applies a preprocessing pipeline before the point cloud is fed into RandLA-Net. This pipeline is centered on slope rotation and flattening, and consists of five stages: voxel downsampling, slope rotation flattening, sliding-window partitioning, intra-block normalization, and fixed-point sampling.

\paragraph{Voxel Downsampling}

The original ground point cloud may contain a large number of points, which increases the computational burden of subsequent processing. Therefore, uniform voxel-grid downsampling is first applied. The space is partitioned into voxel cells with a fixed grid size $g$, and only one point within each voxel is retained. This step significantly compresses the point-cloud scale while preserving spatial uniformity, providing a stable input for subsequent slope rotation and block partitioning.

\paragraph{Slope Rotation and Flattening}

Slope rotation and flattening is the core step of the preprocessing pipeline. Its purpose is to rotate the inclined slope to an approximately horizontal orientation, thereby reducing the interference of the macroscopic elevation trend on network inputs.

First, the principal plane of the current point cloud $p_{i} \in \mathbb{R}^{N \times 3}$ is fitted using the Random Sample Consensus (RANSAC) algorithm \citep{fischler1981ransac}. In each iteration, three non-collinear points  $p_{1}$, $p_{2}$, and $p_{3}$ are randomly sampled from the point cloud, and the unit normal vector of the plane they span is computed as:

\[
n^{\prime} = \frac{( p_{2} - p_{1} ) \times ( p_{3} - p_{1} )}{\left\| ( p_{2} - p_{1} ) \times ( p_{3} - p_{1} ) \right\|}
\]

The number of inliers whose distances to the plane are smaller than the threshold $\tau$ is then counted. After the preset number of iterations, the candidate plane with the maximum number of inliers is selected, and its inliers are further refined by PCA fitting to obtain a robust principal-plane normal vector $n \in \mathbb{R}^{3}$ in the least-squares sense. The sign convention $n_{z} >0$ is enforced to ensure that the normal points upward.

Next, based on the obtained normal vector $n$, the rotation matrix $R$ required to align $n$ with the world-coordinate $+Z$ direction $e_{z} = (0,0,1)^{\top}$ is computed. Let the rotation axis be

$v=n \times e_{z}$, and the sine be $s= \left\| v \right\|$. According to Rodrigues' formula,

\[
R=I+ {[v]}_{\times} + {[v]}_{\times}^{2} \frac{1-c}{s^{2}}
\]

where ${[v]}_{\times}$ is the skew-symmetric matrix of $v= (v_{1}, v_{2}, v_{3})^{\top}$.

\[
{[v]}_{\times} = \left( \begin{matrix}0 & - v_{3} & v_{2} \\ v_{3} & 0 & - v_{1} \\ - v_{2} & v_{1} & 0\end{matrix} \right)
\]

Finally, the point-cloud centroid $\bar{p}$ is computed, the cloud is first translated to the origin, and then the rotation is applied as:

\[
p_i^{\prime} =R \left( p_{i} - \bar{p} \right)
\]

After this transformation, the principal plane of the slope is aligned with the $XOY$ plane. At the same time, the transformation parameters for each point-cloud file (normal vector, rotation matrix, centroid coordinates, and the z-value ranges before and after transformation, etc.) are saved to support inverse transformation of the predicted results back to the original coordinate system.

\paragraph{Sliding-Window Partitioning}

After coordinate transformation, sliding-window partitioning is performed on the flattened point cloud. A square window with side length $L$ slides over the $XOY$ plane with step size $\Delta$ to generate multiple sub-blocks. Sparse sub-blocks with fewer points than the threshold are discarded to ensure sufficient local point density in each training sample.

\paragraph{Intra-Block Normalization}

For each sub-block, the centroid of the point cloud within the block is first subtracted to remove translational offsets. The coordinates are then normalized using half of the block size $L/2$ as the scale factor, compressing all axes into the range $[-1,1]$ and reducing numerical bias introduced by absolute coordinate differences among blocks.

\paragraph{Fixed-Point Sampling}

To satisfy the network's requirement for a fixed input size, each sub-block is finally sampled to a fixed number of points $N_{\mathrm{fix}}$. When the number of points in a block exceeds $N_{\mathrm{fix}}$, random sampling without replacement is performed; when the number is insufficient, sampling with replacement is used to ensure consistency in batch dimensions.

\subsubsection{Model Training}

After preprocessing, each point-cloud sub-block is used as a training sample. The input feature of each point is its three-dimensional coordinate $(x,y,z)$, and the supervision signal is the corresponding point-wise semantic label. The network outputs class probabilities for three categories: normal background, uplift hazard, and subsidence hazard.

Because normal background points greatly outnumber uplift and subsidence hazard points, the training process adopts weighted cross-entropy loss to reduce the influence of class imbalance. Before training, the number of points in each class $n_{c}$ is counted over the training set, and inverse-frequency class weights are computed. To avoid unstable training caused by excessively large weights for rare classes, a smoothing term $\epsilon$ and an upper bound $w_{\max}$ are introduced:

\[
{\tilde{w}}_{c} = \min \left( \frac{1}{\frac{n_{c}}{\sum_{c'} n_{c'}} + \epsilon}, w_{\max} \right)
\]

The normalized class weights are then used in the weighted cross-entropy loss, encouraging the model to pay greater attention to hazard classes while maintaining stable training.

\subsubsection{Full-Point Inference and Label Mapping}

During inference, a single LAS file is used as input. Unlike the training preprocessing, inference is performed on the full point cloud without voxel downsampling, so that the final prediction covers every original point. The same slope-flattening transformation is first applied, while the original coordinates are preserved for final label writing.

The flattened point cloud is partitioned by sliding windows. Each block is normalized, sampled, and fed into the trained model to obtain Softmax probabilities. The probabilities are accumulated in a global probability buffer $S \in \mathbb{R}^{N \times C}$, and a coverage-count array records how many times each point is predicted. The final label is obtained by averaging probabilities and applying:

\[
{\hat{y}}_{i} = \operatorname*{arg\,max}_{c} \frac{S_{i,c}}{{\mathrm{cnt}}_{i}}
\]

This probability accumulation reduces prediction variance in overlapping regions.

For boundary points not covered by any window, a KD-Tree is used to find the nearest covered point, and its probability vector is assigned to the uncovered point. After full-point prediction is completed, the predicted semantic labels are written back to the original LAS file for visualization and analysis.

\begin{figure}[pos=H]
  \centering
  \includegraphics[width=\figwidth,height=.68\textheight,keepaspectratio]{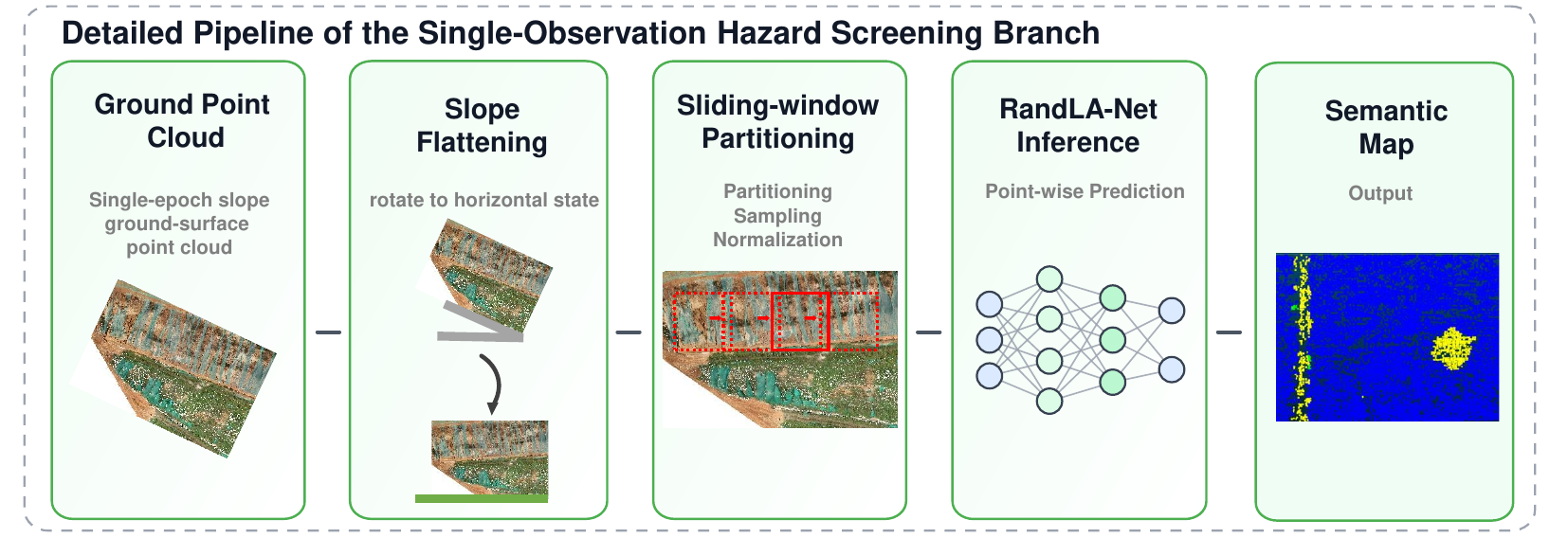}
  \caption{Detailed pipeline of the single-observation hazard screening branch, including ground-point input, slope flattening, sliding-window partitioning, RandLA-Net inference, and semantic map output.}\label{fig:6}
\end{figure}

\subsection{Multi-Epoch Deformation Monitoring Branch}

\subsubsection{Functional Role and Input--Output Definition}

The multi-epoch deformation monitoring branch is responsible for observing temporal elevation changes on the slope surface. Its inputs are ground-surface point clouds extracted from two time-separated UAV-borne LiDAR survey epochs. After point-cloud registration, gridding, common-grid intersection, and elevation differencing, this branch outputs an elevation-difference heatmap in the original point-cloud coordinate system, thereby providing an intuitive representation of slow surface change.

\subsubsection{Inter-Epoch Point-Cloud Registration}

Point clouds acquired at different epochs may contain systematic rigid offsets in the global coordinate system because of positioning errors, flight attitude differences, atmospheric conditions, or trajectory reconstruction uncertainty. If direct differencing is performed on unregistered point clouds, these coordinate discrepancies may be falsely interpreted as actual ground deformation signals. Therefore, registration is performed before differencing to estimate and reduce the rigid transformation deviation between epochs.

For a stable ground location, the two reconstructed epochs may be approximated as

\[
p^{(2)} \approx R p^{(1)}+t+\epsilon,
\]

where $R$ and $t$ denote residual inter-epoch rotation and translation and $\epsilon$ denotes measurement noise. This relation shows that an apparent coordinate difference may come from acquisition bias rather than true deformation, which is why registration is treated as the first step of the multi-epoch branch.

If two epochs are differenced before registration, the apparent residual contains both true change and rigid acquisition bias:

\[
p^{(2)}-p^{(1)} \approx (R-I)p^{(1)}+t+\epsilon.
\]

For centimeter-level deformation monitoring, this residual may be comparable to the actual elevation change and must be reduced before quantitative differencing.

Based on the preceding preprocessing stage, ground-surface point clouds from two epochs are used as the registration targets. This study adopts the Point-to-Point Iterative Closest Point (ICP) algorithm \citep{besl1992icp} to estimate the rotation matrix and translation vector between the two point clouds. Since both UAV surveys are conducted under the same projected coordinate system, the initial misalignment is usually within a limited range, allowing ICP to converge reliably.

Let the reference ground point cloud be $P_{1} = {\left\{ p_{i}^{\left( 1 \right)} \right\}}_{i=1}^{M}$ and the target ground point cloud be $P_{2} = {\left\{ p_{j}^{\left( 2 \right)} \right\}}_{j=1}^{N}$. The optimization objective of the Point-to-Point ICP is to estimate the rotation matrix $R \in \mathrm{SO}(3)$ and translation vector $t \in \mathbb{R}^{3}$, such that the sum of squared Euclidean distances between corresponding point pairs is minimized:

\[
\left( \hat{R}, \hat{t} \right) = \operatorname*{arg\,min}_{R \in \mathrm{SO}(3), t \in \mathbb{R}^{3}} \sum_{i=1}^{M} {\left\| R p_{i}^{\left( 1 \right)} +t- q_{i} \right\|}^{2}
\]

where $q_{i} = \operatorname*{arg\,min}_{p \in P_{2}} \left\| p- p_{i}^{\left( 1 \right)} \right\|$ denotes the closest point in $P_{2}$ to $p_{i}^{\left( 1 \right)}$):

The algorithm iteratively alternates between nearest neighbor correspondence search and transformation estimation. In each iteration, for every point in the source point cloud $P_{1}$, the nearest neighbor in the target point cloud $P_{2}$ is identified to form a set of corresponding point pairs $\left\{ \left( p_{i}^{\left( 1 \right)}, q_{i} \right) \right\}$. Then, the optimal rotation matrix $R$ and translation vector $t$ for the current correspondences are solved using Singular Value Decomposition (SVD). The iteration continues until the change in Root Mean Square Error (RMSE) between successive iterations falls below a predefined threshold:

\[
\mathrm{RMSE}^{\left( k \right)} = \sqrt{\frac{1}{M} \sum_{i=1}^{M} {\left\| R^{\left( k \right)} p_{i}^{\left( 1 \right)} + t^{\left( k \right)} - q_{i}^{\left( k \right)} \right\|}^{2}} < \tau
\]

Through the above iterative process, the aligned ground point clouds of the two epochs are obtained, providing a consistent spatial reference for subsequent grid-based differential analysis.

\subsubsection{Grid Construction for Surface Statistics}

After point cloud registration is completed, the two epochs of ground point clouds must be transformed into structured spatial statistical units so that elevation statistics can be computed on a region-by-region basis and inter-epoch differencing can be performed. This step is necessary because independent LiDAR flights do not repeatedly sample the exact same physical ground points. In time-of-flight LiDAR, the range of each return is computed from pulse travel time,

\[
d=\frac{c\Delta t}{2},
\]

and the final point location also depends on the beam direction at the scanning instant. Therefore, two flights over the same slope generally produce different discrete samples of the same continuous surface; even after registration, nearest-neighbor pairs may still correspond to slightly different surface locations. This study therefore adopts a uniform rectangular gridding scheme based on real geographic coordinates. The horizontal extent covered by the ground point clouds is partitioned into regular two-dimensional grids with a fixed step size $s$ (i.e., the grid side length, measured in meters, with the specific value determined in the experimental section of Section~\ref{sec:experiments}), and each grid cell serves as the fundamental spatial unit for subsequent statistical analysis and differencing.

Specifically, let the horizontal extent of the ground point cloud in the projected coordinate system be $\left[ X_{min}, X_{max} \right] \times \left[ Y_{min}, Y_{max} \right]$. Then, the grid-cell indices $\left( g_{i}, g_{j} \right)$ to which any point $p= \left( x,y,z \right)$ belongs are determined as follows:

\[
g_{i} = \left\lfloor \frac{x- X_{min}}{s} \right\rfloor, g_{j} = \left\lfloor \frac{y- Y_{min}}{s} \right\rfloor
\]

Each point is uniquely assigned to the corresponding grid cell according to its horizontal coordinates. The grid key is defined as an integer tuple $\left( g_{i}, g_{j} \right)$ and stored in the form of a hash dictionary, where each dictionary value is the set of indices of all points falling within that grid cell. This structure enables direct access to the point set of any grid cell in $O(1)$ time, thereby providing efficient data support for large-scale batch statistical computation. The above gridding procedure is applied independently to the two epochs of ground point clouds, yielding the baseline-epoch grid index dictionary $G_{1}$ and the second-epoch grid index dictionary $G_{2}$.

\begin{algorithm}[htbp]
\caption{Grid index construction for surface statistics}
\label{alg:grid-indexing}
\begin{algorithmic}[1]
\Require Ground-point set $P=\{p_i=(x_i,y_i,z_i)\}_{i=1}^{N}$; grid size $s$
\Ensure Grid-index dictionary $G$
\State $X_{\min} \gets \min_i x_i$, $Y_{\min} \gets \min_i y_i$
\State Initialize $G \gets \emptyset$
\For{each point $p_i=(x_i,y_i,z_i)\in P$}
  \State $g_x \gets \left\lfloor (x_i-X_{\min})/s \right\rfloor$
  \State $g_y \gets \left\lfloor (y_i-Y_{\min})/s \right\rfloor$
  \State $k \gets (g_x,g_y)$
  \If{$k \notin G$}
    \State $G[k] \gets [\,]$
  \EndIf
  \State append point index $i$ to $G[k]$
\EndFor
\State \Return $G$
\end{algorithmic}
\end{algorithm}

\subsubsection{Common-Grid Intersection between Epochs}

Because the two point clouds are acquired independently at different times, their spatial coverage over ground points is inconsistent across epochs. On the one hand, vegetation may grow or die back between the two acquisitions, leading to changes in canopy density. When vegetation is dense, laser pulses have difficulty penetrating the canopy to reach the ground, resulting in sparse ground points or even voids in the corresponding areas; when vegetation is sparse or has died back, the same location may yield more complete ground-point coverage. On the other hand, due to factors such as system error, scanning angle, and non-uniform point density distribution, the two epochs may also exhibit inconsistent point presence or absence in local areas.

These factors jointly cause some grid cells to contain valid points in only one epoch while being empty in the other. If differencing were performed directly on such grid cells, missing data would lead to spurious deformation results. To address this issue, this study takes the intersection of the grid-key sets from the two epochs and retains only those grid cells that have valid ground-point coverage in both epochs for subsequent analysis:

\[
K_{\mathrm{common}} = K^{(1)} \cap K^{(2)}
\]

where $K^{(1)}$ and $K^{(2)}$ denote the grid-key sets of the ground point clouds from the baseline epoch and the epoch to be registered, respectively.

Furthermore, to support batch computation of subsequent differencing vectors, a deterministic linear ordering must be established for $K_{\mathrm{common}}$. In this study, sorting is performed according to the spatial scan order of the second-epoch grid keys, yielding an ordered sequence of valid grids:

\[
K_{\mathrm{common}}^{\mathrm{ordered}} = \left[ k \mid k \in K_{\mathrm{ordered}}^{(2)}, k \in K_{\mathrm{common}} \right]
\]

where $K_{\mathrm{ordered}}^{(2)}$ is the ordered list of second-epoch grid keys arranged according to the spatial scan order. This procedure ensures a strict one-to-one correspondence between the two epochs in the index dimension during subsequent grid-wise statistic extraction and differencing-vector construction, thereby avoiding misaligned calculations caused by the unordered nature of the key sets.

After the intersection operation described above, the final output $K_{\mathrm{common}}^{\mathrm{ordered}}$ contains only those spatial units that are reliably covered by the ground point clouds in both epochs, thereby providing strict data-quality assurance for subsequent within-grid statistic computation and inter-epoch differencing analysis.

\begin{algorithm}[htbp]
\caption{Ordered common-grid intersection between two epochs}
\label{alg:common-grid}
\begin{algorithmic}[1]
\Require Baseline-epoch grid dictionary $G_A$; target-epoch grid dictionary $G_B$
\Ensure Ordered common-grid key list $K_{\mathrm{common}}^{\mathrm{ordered}}$
\State $K_A \gets \mathrm{keys}(G_A)$
\State $K_B \gets \mathrm{keys}(G_B)$
\State $K_{\mathrm{common}} \gets K_A \cap K_B$
\If{$K_{\mathrm{common}} = \emptyset$}
  \State \textbf{raise} an error indicating that no valid common grid cells are available
\EndIf
\State $K_B^{\mathrm{ordered}} \gets$ keys of $G_B$ sorted by target-epoch spatial scan order
\State Initialize $K_{\mathrm{common}}^{\mathrm{ordered}} \gets [\,]$
\For{each key $k \in K_B^{\mathrm{ordered}}$}
  \If{$k \in K_{\mathrm{common}}$}
    \State append $k$ to $K_{\mathrm{common}}^{\mathrm{ordered}}$
  \EndIf
\EndFor
\State \Return $K_{\mathrm{common}}^{\mathrm{ordered}}$
\end{algorithmic}
\end{algorithm}

\subsubsection{Grid-Wise Elevation Statistics and Differencing}

After obtaining the common valid grid set $K_{\mathrm{common}}^{\mathrm{ordered}}$ from the two epochs of ground point clouds, for each grid cell $\left( g_{i}, g_{j} \right)$, the set of points falling into that cell is extracted from the grid index dictionaries of the two epochs, and a set of elevation statistics representing the surface morphology of that grid cell is computed.

Let the set of ground-point elevation values contained in grid cell $\left( g_{i}, g_{j} \right)$ in the first epoch (denoted as $A$) be $Z_{i,j}^{A} = \left\{ z_{1}^{A}, z_{2}^{A}, \ldots, z_{n_{A}}^{A} \right\}$, and let the set of ground-point elevation values contained in the second epoch (denoted as $B$) be $Z_{i,j}^{B} = \left\{ z_{1}^{B}, z_{2}^{B}, \ldots, z_{n_{B}}^{B} \right\}$. The following statistic is computed for each grid cell in this study:

Mean elevation: the arithmetic mean of all ground-point elevations within the grid, used as a representative estimate of the surface elevation of that grid cell:

\[
{\bar{Z}}_{i,j} = \frac{1}{n} \sum_{k=1}^{n} z_{k}
\]

The above statistic is computed independently for the two epochs of ground point clouds, yielding the baseline-epoch mean elevation ${\bar{Z}}_{i,j}^{A}$ and the registered-epoch mean elevation ${\bar{Z}}_{i,j}^{B}$ for each grid cell.

After obtaining the elevation statistics for each grid cell in the two epochs, the inter-epoch elevation difference is computed for each grid cell $\left( g_{i}, g_{j} \right)$ in the common valid grid set. The differencing direction is defined as epoch $B$ minus epoch $A$:

\[
\Delta Z_{i,j} = {\bar{Z}}_{i,j}^{B} - {\bar{Z}}_{i,j}^{A}
\]

where $\Delta Z_{i,j} >0$ indicates an increase in ground elevation within the grid area, which may correspond to material accumulation or surface uplift on the slope;

$\Delta Z_{i,j} <0$ indicates a decrease in ground elevation, which may correspond to deformation behaviors such as slope sliding, surface erosion, or settlement;

$\Delta Z_{i,j} \approx 0$ indicates that the surface elevation remained stable in the observed period.

Applying the above differencing operation to all $K$ common valid grids yields an elevation-difference vector $\Delta Z \in \mathbb{R}^{K}$ covering the entire slope under investigation. Its spatial resolution is determined by the grid step size $s$, and each element corresponds to the inter-epoch change in the mean elevation of ground points within an $s \times s$ ground area.

\subsubsection{Elevation-Difference Heatmap Generation}

To convert the one-dimensional difference vector into a two-dimensional image with spatial readability, this study constructs a mapping from grid keys to matrix row and column indices for all grid keys $\left( g_{i}, g_{j} \right)$ in  $K_{\mathrm{common}}^{\mathrm{ordered}}$:

\[
g_{i} \mapsto i_{x} \in \{0,1, \ldots, N_{x} -1\}, g_{j} \mapsto j_{y} \in \{0,1, \ldots, N_{y} -1\}
\]

where $N_{x}$ and $N_{y}$ are the numbers of unique coordinates of the common grids along the $X$ and $Y$ directions, respectively. A zero matrix of size $N_{y} \times N_{x}$ is then initialized to obtain the elevation-difference heatmap matrix $H_{\Delta Z} \in \mathbb{R}^{N_{y} \times N_{x}}$, where each element spatially corresponds to an $s \times s$ ground area. Finally, the difference heatmap is overlaid semi-transparently on the orthographic projection image of the point cloud, so that the spatial location of deformation areas can be directly associated with the actual ground features, facilitating timely localization of potential hazard zones by relevant personnel.

\begin{figure}[pos=H]
  \centering
  \includegraphics[width=\figwidth,height=.68\textheight,keepaspectratio]{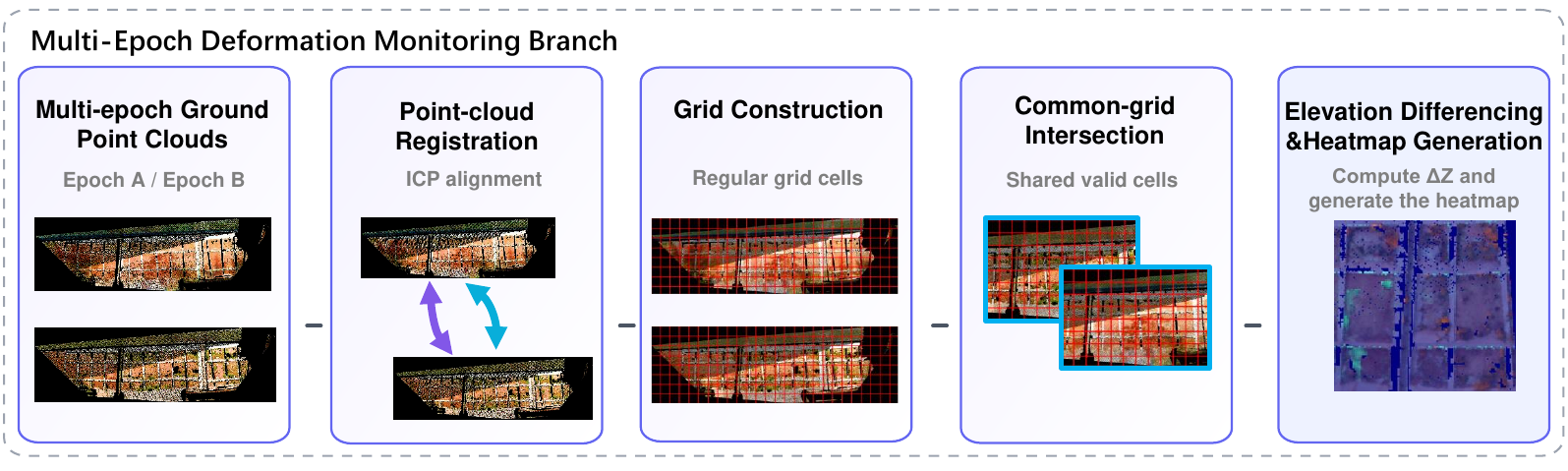}
  \caption{Detailed pipeline of the multi-epoch deformation monitoring branch, including point-cloud registration, grid construction, common-grid intersection, elevation differencing, and heatmap generation.}\label{fig:7}
\end{figure}

\section{Experiments}\label{sec:experiments}

This section evaluates the proposed workflow through three linked experiments. The first experiment verifies whether the shared CSF-based preprocessing stage can provide usable ground-surface point clouds under vegetation cover. The second experiment evaluates the single-observation hazard-screening branch, which must identify local geometric anomalies without a historical baseline. The third experiment evaluates the multi-epoch deformation-monitoring branch, which must quantify temporal elevation changes from time-separated UAV-borne LiDAR observations. To make the experimental evidence self-contained, the main parameter settings, dataset statistics, visual results, and quantitative indicators are reported together in this section.

\subsection{Ground-Surface Extraction Validation}\label{sec:exp-ground-extraction}

The under-canopy foam-board experiment was used to verify whether UAV-borne LiDAR combined with CSF filtering can retain usable ground-surface information under vegetation occlusion. A known low-relief foam-board target was placed beneath canopy cover, and the original mixed point cloud was compared with the CSF-filtered ground points. The CSF parameters used in this experiment are listed in Table~\ref{tbl:csf-params}.

\begin{table}[width=.95\linewidth,cols=3,pos=htbp]
\caption{CSF parameter settings for ground-surface extraction}\label{tbl:csf-params}
\begin{tabular*}{\tblwidth}{@{} LLL @{}}
\toprule
Parameter & Value & Description \\
\midrule
cloth resolution & 0.1 m & Cloth grid size \\
rigidness & 1 & Terrain adaptability \\
time step & 0.5 & Deformation step \\
class threshold & 0.05 m & Ground threshold \\
\bottomrule
\end{tabular*}
\end{table}

\begin{figure}[pos=H]
  \centering
  \includegraphics[width=\figwidth,height=.68\textheight,keepaspectratio]{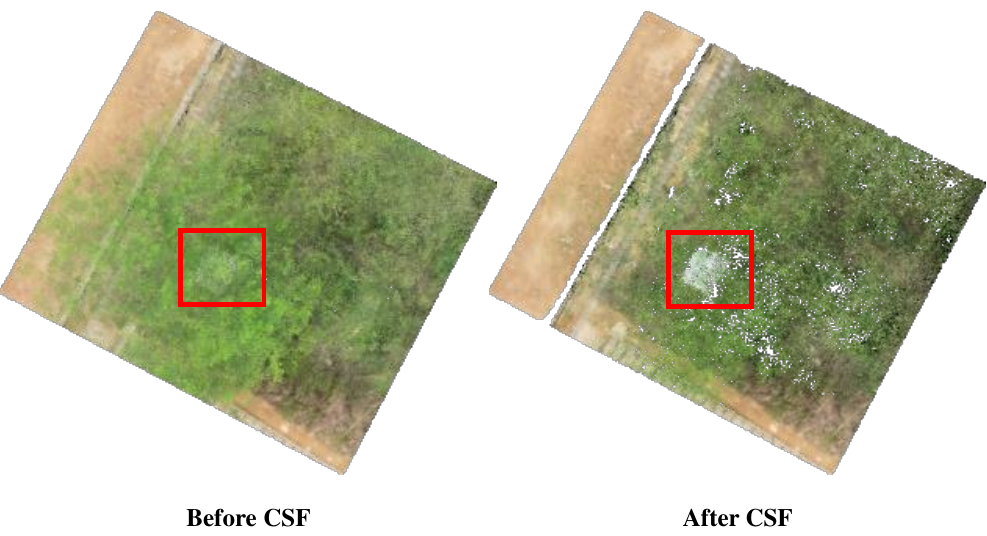}
  \caption{Visual comparison of the under-canopy foam-board target before and after CSF filtering.}\label{fig:8}
\end{figure}

Figure~\ref{fig:8} shows the visual comparison before and after CSF filtering. Before filtering, the point cloud contains a mixture of canopy, branches, surrounding objects, ground points, and the foam-board target. After CSF filtering, many high-elevation non-ground returns are removed, while the foam-board target and nearby ground-surface points remain visible in the filtered point set.

\begin{figure}[pos=H]
  \centering
  \includegraphics[width=\figwidth,height=.68\textheight,keepaspectratio]{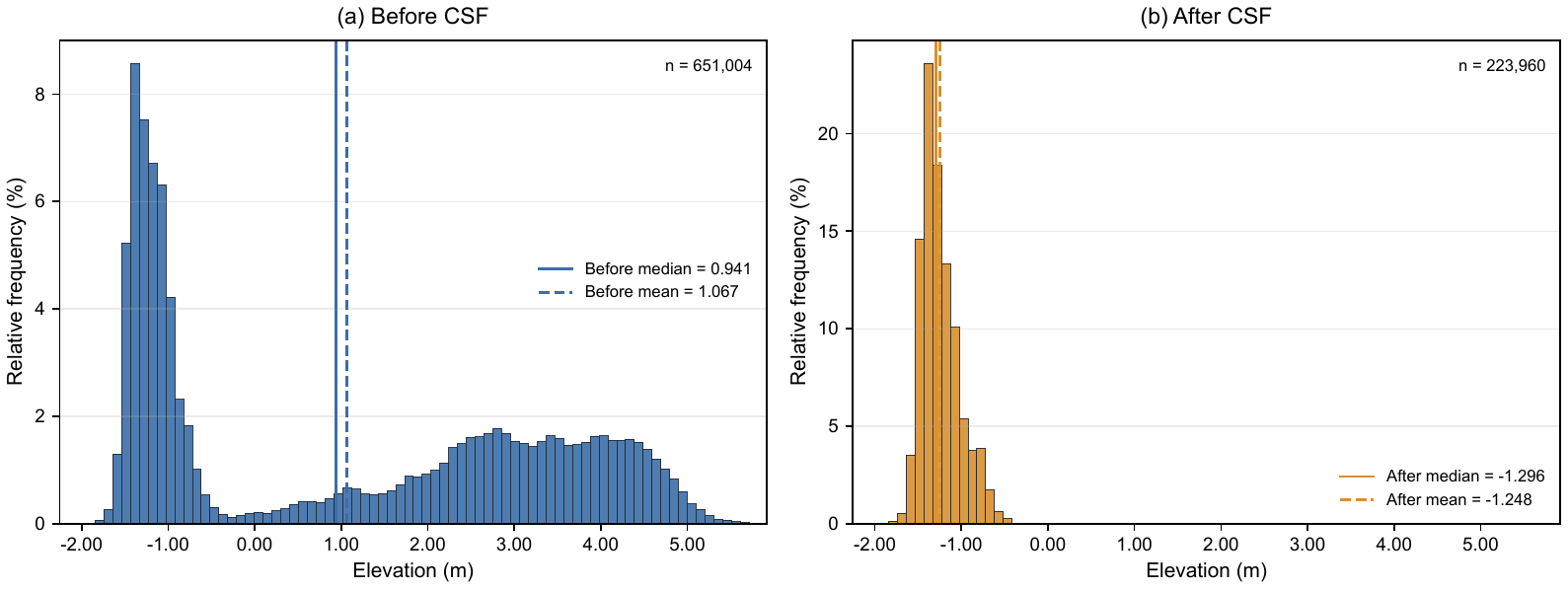}
  \caption{Elevation histograms of the foam-board ROI before and after CSF filtering, illustrating the reduction of high-elevation non-ground returns and the concentration of filtered ground points in the low-elevation range.}\label{fig:9}
\end{figure}

The histogram in Fig.~\ref{fig:9} provides a statistical view of the filtering effect. Before CSF processing, the elevation distribution contains both low-elevation returns from the foam-board target and surrounding ground surface, as well as high-elevation returns from canopy, branches, and other above-ground objects. After CSF processing, the distribution becomes more concentrated in the low-elevation range, while the high-elevation tail is clearly reduced. This change indicates that CSF suppresses a large proportion of non-ground returns without completely removing the low-relief target and adjacent ground points.

This result is important for the workflow because both downstream branches depend on the quality of the extracted ground-surface point cloud. For the single-observation branch, a cleaner ground representation helps the network focus on local geometric anomalies instead of canopy clutter. For the multi-epoch branch, reducing above-ground returns decreases the risk that vegetation differences between epochs will be interpreted as terrain deformation.

\subsection{Single-Observation Hazard Screening}\label{sec:exp-single-observation}

The single-observation branch was evaluated using simulated slope-hazard point clouds with three semantic classes: normal background, uplift hazard, and subsidence hazard. The purpose was to test whether the RandLA-Net-based branch can localize potential hazard regions from one ground-surface point cloud when no historical baseline is available. The simulated dataset contained 92 slope scenes, and its class distribution is summarized in Table~\ref{tbl:class-distribution}.

\begin{table}[width=.95\linewidth,cols=3,pos=htbp]
\caption{Class distribution statistics of the simulated dataset}\label{tbl:class-distribution}
\begin{tabular*}{\tblwidth}{@{} LLL @{}}
\toprule
Class & Total points & Proportion (\%) \\
\midrule
Normal background & 47,406,226 & 54.80 \\
Bulging hazard & 19,537,733 & 22.59 \\
Depression hazard & 19,556,338 & 22.61 \\
Total & 86,500,297 & 100 \\
\bottomrule
\end{tabular*}
\end{table}

The preprocessing and training settings used for the single-observation branch are summarized in Table~\ref{tbl:preprocess-settings}. These settings correspond to voxel downsampling, slope rotation and flattening, sliding-window partitioning, sparse-block filtering, and fixed-size sampling.

\begin{table}[width=.95\linewidth,cols=3,pos=htbp]
\caption{Data preprocessing hyperparameter settings}\label{tbl:preprocess-settings}
\begin{tabular*}{\tblwidth}{@{} LLL @{}}
\toprule
Hyperparameter & Setting & Description \\
\midrule
Voxel size & 0.01 m & Downsampling grid \\
Plane fitting & RANSAC & Main-plane estimation \\
RANSAC iterations & 64 & Fitting iterations \\
Inlier threshold & 0.08 m & Inlier distance \\
Block size & 20 m & Window size \\
Sliding step & 18.0 m & 2 m overlap \\
Minimum points & 4096 & Sparse-block filter \\
Sample size & 131072 & Points per block \\
Train/validation split & 8:2 & Data split \\
\bottomrule
\end{tabular*}
\end{table}

Scene-level 5-fold cross-validation was used so that the scenes used for validation were not included in training. As shown in Table~\ref{tbl:5-5}, the model achieved a mean overall accuracy (OA) of 64.71\% and a mean mIoU of 46.48\%. Across the five folds, OA ranged from 62.18\% to 67.98\%, and mIoU ranged from 43.87\% to 50.65\%. The moderate inter-fold variation indicates that the model learned repeatable geometric patterns from the simulated scenes, although the performance is still preliminary for engineering deployment.

\begin{table}[width=\linewidth,cols=7,pos=htbp]
\caption{Scene-level 5-fold cross-validation results for the single-observation hazard-screening branch}\label{tbl:5-5}
\footnotesize
\begin{tabular*}{\tblwidth}{@{} LLLLLLL @{} }
\toprule
Fold & Train./val. scenes & OA (\%) & mIoU (\%) & Bg IoU (\%) & Bulge IoU (\%) & Dep. IoU (\%) \\
\midrule
Fold 1 & 73/19 & 67.98 & 50.65 & 55.12 & 47.26 & 49.58 \\
Fold 2 & 73/19 & 62.18 & 43.87 & 50.74 & 40.97 & 39.92 \\
Fold 3 & 74/18 & 64.80 & 46.06 & 54.06 & 41.91 & 42.21 \\
Fold 4 & 74/18 & 65.35 & 46.63 & 54.97 & 41.98 & 42.95 \\
Fold 5 & 74/18 & 63.24 & 45.19 & 50.59 & 42.18 & 42.80 \\
Mean & 73.6/18.4 & 64.71 & 46.48 & 53.10 & 42.86 & 43.49 \\
Std. & -- & 2.22 & 2.55 & 2.26 & 2.51 & 3.61 \\
\bottomrule
\end{tabular*}
\end{table}

\begin{figure}[pos=H]
  \centering
  \includegraphics[width=\figwidth,height=.68\textheight,keepaspectratio]{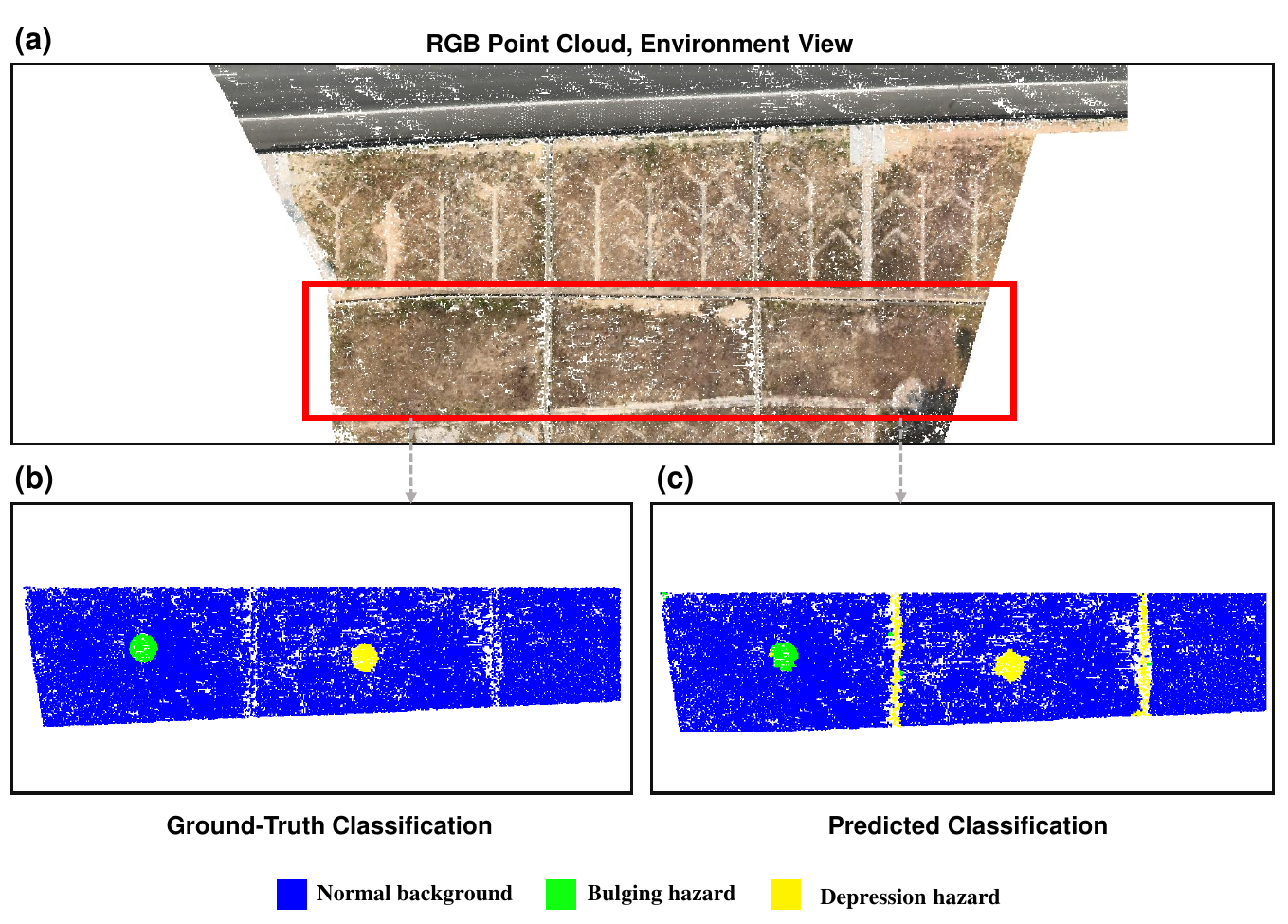}
  \caption{Qualitative result of the single-observation hazard screening branch, comparing the RGB point-cloud scene, ground-truth labels, and predicted labels for normal background, bulging hazard, and depression hazard.}\label{fig:10}
\end{figure}

The qualitative result in Fig.~\ref{fig:10} further shows that the predicted labels can locate the main uplift and depression regions in the test scene, despite boundary errors and local misclassification. Two drainage-channel areas are incorrectly rendered in the depression-hazard color because their elongated low-relief geometry resembles the simulated depression class. This false-positive pattern suggests that future training data should include more drainage-channel and engineered-structure samples, but it does not alter the main conclusion that the branch can provide preliminary no-baseline hazard localization.

\subsection{Multi-Epoch Deformation Monitoring}\label{sec:exp-multi-epoch}

The multi-epoch branch was validated using both a controlled known-deformation experiment and a real expressway slope case. These two experiments serve complementary purposes. The controlled experiment tests whether the gridded differencing method can recover known artificial elevation changes, whereas the real-field case tests whether the same workflow can localize progressive surface change under practical slope conditions. The main UAV-borne LiDAR acquisition settings used for the two experiments are listed in Table~\ref{tbl:lidar-params}.

\begin{table}[width=.95\linewidth,cols=2,pos=htbp]
\caption{Main UAV-borne LiDAR acquisition parameters}\label{tbl:lidar-params}
\begin{tabular*}{\tblwidth}{@{} LL @{}}
\toprule
Parameter & Value \\
\midrule
Flight altitude & 40 m \\
Flight speed & 3 m/s \\
Sampling frequency & 240 kHz \\
Scanning mode & Non-repetitive \\
Lateral laser overlap & 50\% \\
Number of returns & Five returns \\
\bottomrule
\end{tabular*}
\end{table}

For the controlled experiment, two artificial uplift regions were constructed beneath vegetation cover. The square flat-topped uplift had a measured height of approximately 16 cm, and the gridded differencing result produced a mean elevation difference of 13.31 cm over the delineated region. Because this uplift was designed as an approximately uniform block-like elevation increase, the mean value is the appropriate representative statistic. The circular conical uplift had a measured peak height of approximately 36 cm, and the maximum grid-wise differencing response was 33.9 cm. Because this uplift decreases from the center toward the boundary, the maximum grid-wise value is more suitable than the regional mean for evaluating the peak response. The quantitative results are summarized in Table~\ref{tbl:5-7}.

\begin{figure}[pos=H]
  \centering
  \includegraphics[width=\figwidth,height=.68\textheight,keepaspectratio]{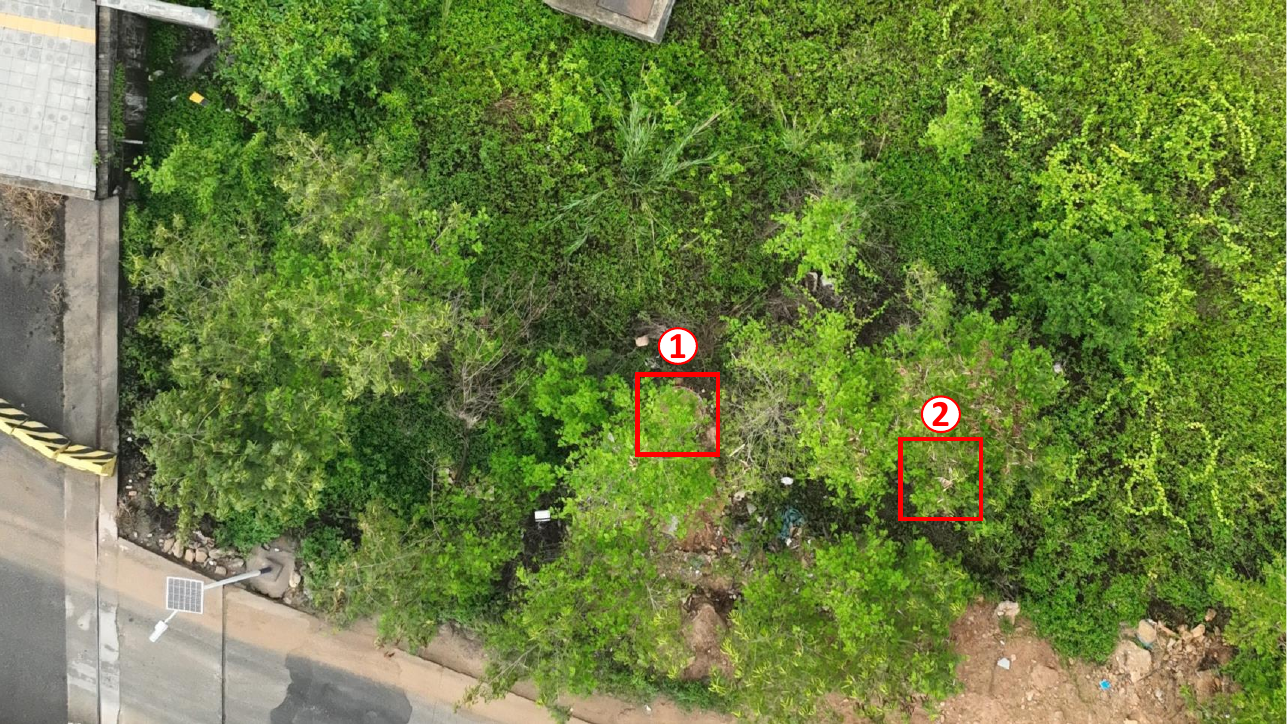}
  \caption{Aerial overview of the controlled known-deformation experiment site, with two manually constructed artificial uplift regions marked for multi-epoch elevation-differencing validation.}\label{fig:11}
\end{figure}

Figure~\ref{fig:11} shows the controlled known-deformation experiment site and the two manually constructed regions. The manually constructed surfaces were not perfectly regular because nearby natural soil was used in field conditions; therefore, small boundary and surface irregularities were retained in the experiment.

\begin{table}[width=.95\linewidth,cols=3,pos=htbp]
\caption{Core quantitative differencing results for the controlled known-deformation experiment}\label{tbl:5-7}
\begin{tabular*}{\tblwidth}{@{} LLL @{} }
\toprule
Region & Field measurement & Representative differencing result \\
\midrule
Square flat-topped uplift & Side length $\approx$ 1.0 m; height $\approx$ 16 cm & Mean $\Delta Z = 13.31$ cm \\
Circular conical uplift & Diameter $\approx$ 1.06 m; peak height $\approx$ 36 cm & Maximum $\Delta Z = 33.9$ cm \\
\bottomrule
\end{tabular*}
\end{table}

\begin{figure}[pos=H]
  \centering
  \includegraphics[width=\figwidth,height=.68\textheight,keepaspectratio]{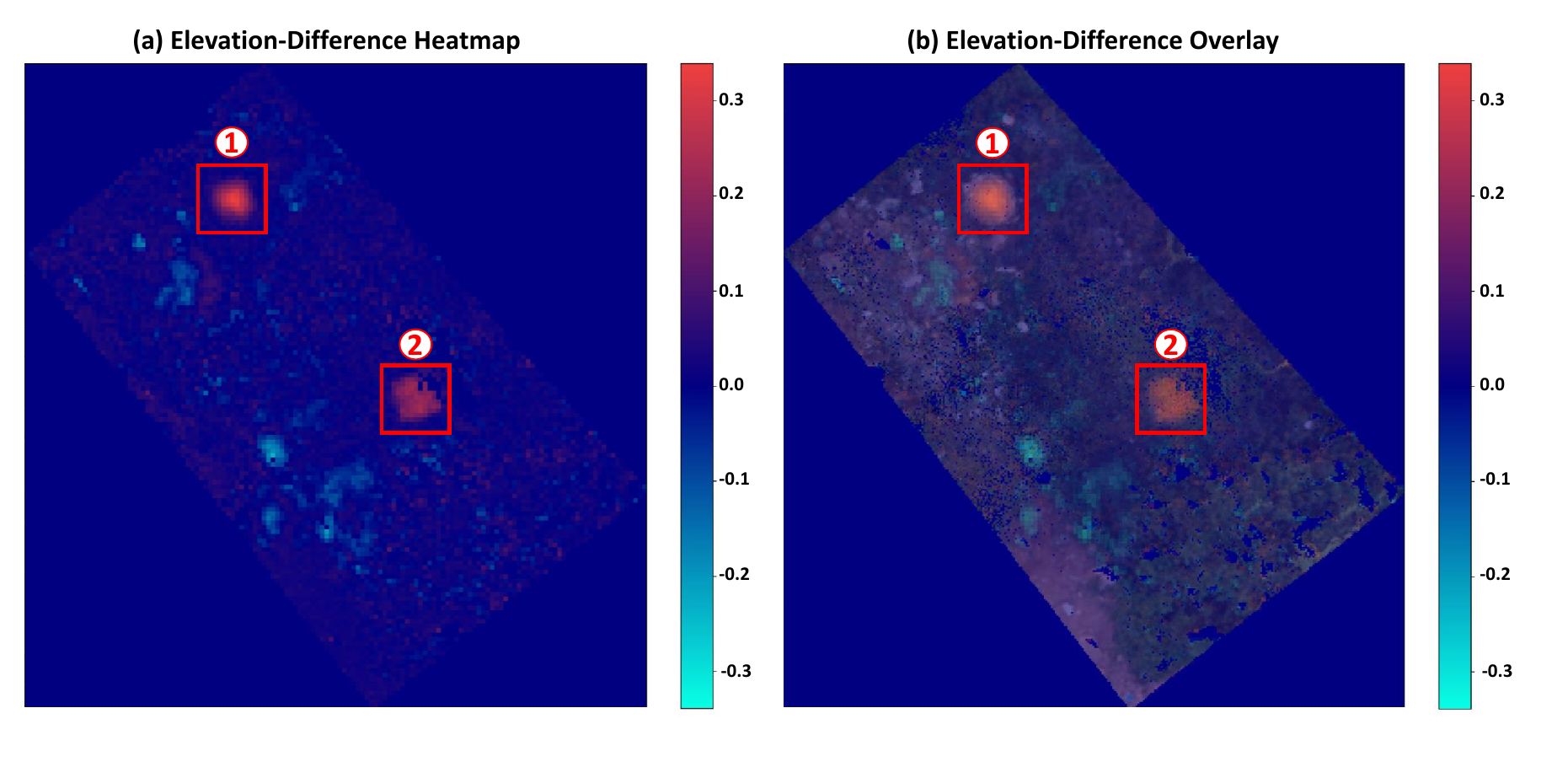}
  \caption{Elevation-difference results of the controlled known-deformation experiment, showing the spatial responses of the two artificial uplift regions in both heatmap and overlay views.}\label{fig:12}
\end{figure}

Figure~\ref{fig:12} shows the corresponding elevation-difference heatmap and overlay view. The two artificial uplift regions appear as concentrated positive elevation-change responses in both views, rather than as isolated noisy pixels. This spatial continuity supports the quantitative interpretation in Table~\ref{tbl:5-7} and confirms that the branch can detect local elevation increases under controlled field conditions.

\begin{figure}[pos=H]
  \centering
  \includegraphics[width=\figwidth,height=.68\textheight,keepaspectratio]{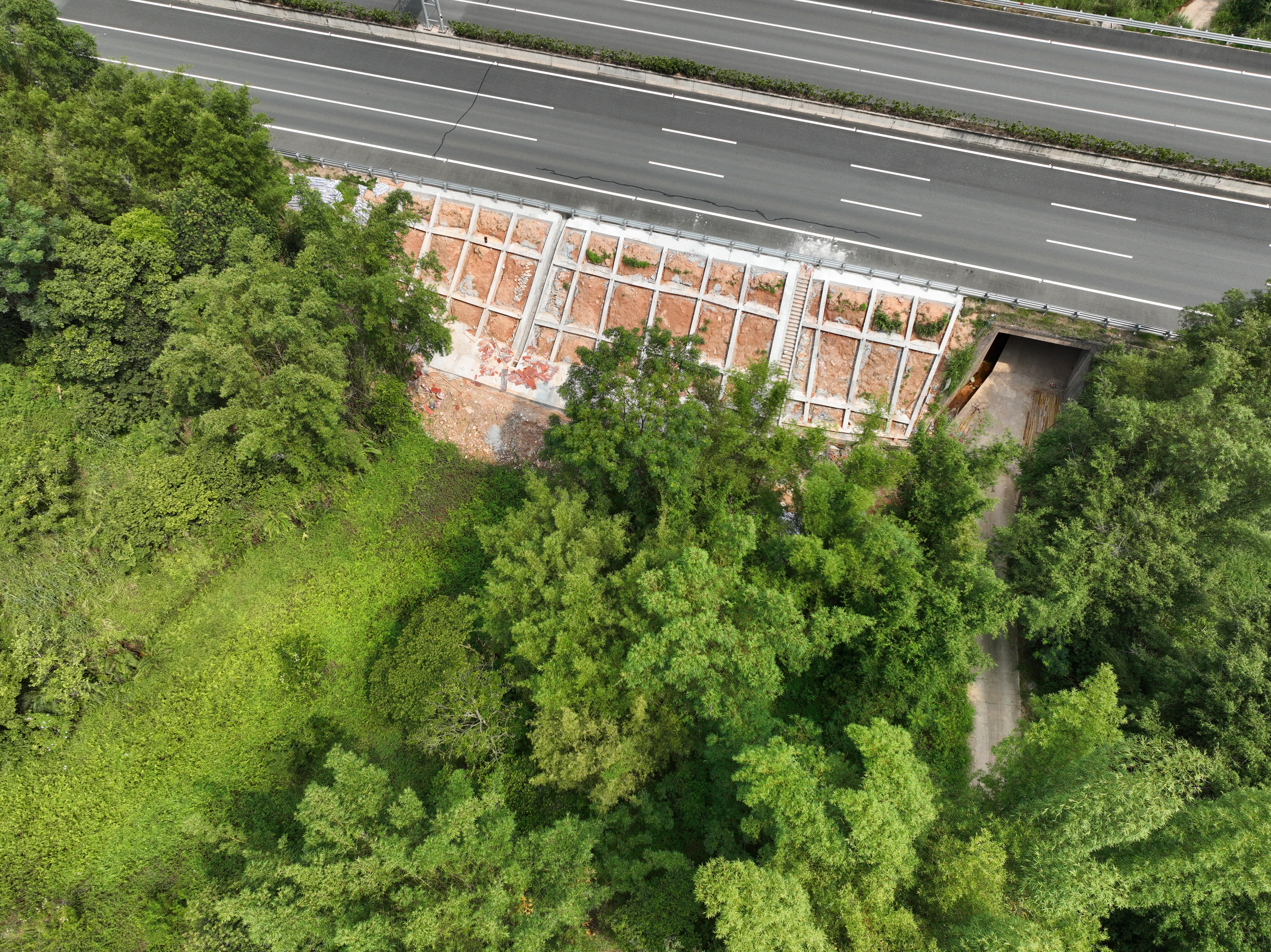}
  \caption{Aerial overview of the real expressway slope field site.}\label{fig:13}
\end{figure}

The real expressway slope experiment used two UAV-borne LiDAR observations acquired on 15 Dec 2025 and 25 Mar 2026. Figure~\ref{fig:13} shows the field site, including the vegetated slope area, roadway, and exposed lattice-beam slope surface.

The basic raw point-cloud statistics of the two epochs are listed in Table~\ref{tbl:raw-stats}. The two epochs had similar coverage area and point density, providing a comparable basis for multi-epoch differencing.

\begin{table}[width=.95\linewidth,cols=3,pos=htbp]
\caption{Basic statistics of raw point clouds from the two epochs}\label{tbl:raw-stats}
\begin{tabular*}{\tblwidth}{@{} LLL @{}}
\toprule
Statistic & Epoch 1 & Epoch 2 \\
\midrule
Number of points & 1,021,818 & 1,070,296 \\
Coverage area & 2,767.98 m$^2$ & 2,703.34 m$^2$ \\
Point density & 369.15 points/m$^2$ & 395.91 points/m$^2$ \\
\bottomrule
\end{tabular*}
\end{table}

After ROI clipping, the retained point clouds were limited to the monitored slope area. The corresponding statistics are listed in Table~\ref{tbl:roi-stats}. CSF filtering was then applied to extract ground-surface points, and the extraction results are summarized in Table~\ref{tbl:ground-stats}.

\begin{table}[width=.95\linewidth,cols=3,pos=htbp]
\caption{Point-cloud statistics after ROI clipping}\label{tbl:roi-stats}
\begin{tabular*}{\tblwidth}{@{} LLL @{}}
\toprule
Statistic & Epoch 1 & Epoch 2 \\
\midrule
Number of ROI points & 584,523 & 600,768 \\
ROI area & 1,799.70 m$^2$ & 1,809.96 m$^2$ \\
ROI point density & 324.79 points/m$^2$ & 331.92 points/m$^2$ \\
\bottomrule
\end{tabular*}
\end{table}

\begin{table}[width=.95\linewidth,cols=3,pos=htbp]
\caption{Statistics of CSF ground-point extraction results}\label{tbl:ground-stats}
\begin{tabular*}{\tblwidth}{@{} LLL @{}}
\toprule
Statistic & Epoch 1 & Epoch 2 \\
\midrule
Ground points after CSF & 115,221 & 162,125 \\
Ground-point ratio & 19.71\% & 26.99\% \\
\bottomrule
\end{tabular*}
\end{table}

Using the first-epoch ground-surface point cloud as the reference, Point-to-Point ICP registration was performed on the second-epoch ground-surface point cloud. The maximum correspondence distance was set to 0.10 m, and the maximum number of iterations was set to 200. The ICP registration achieved an inlier RMSE of 0.025999 m. The overall nearest-neighbor RMSE decreased from 0.036416 m before registration to 0.034315 m after registration, corresponding to an improvement of 5.77\%.

After registration, the two ground-surface point clouds were gridded and intersected on common valid cells. The gridding and common-grid statistics are listed in Table~\ref{tbl:grid-stats}. The workflow achieved a common grid coverage rate of 94.71\%, indicating that most valid grid cells retained paired elevation information for differencing.

\begin{table}[width=.95\linewidth,cols=4,pos=htbp]
\caption{Statistics of gridding and common-grid intersection}\label{tbl:grid-stats}
\begin{tabular*}{\tblwidth}{@{} LLLL @{}}
\toprule
Grid step & Epoch 1 valid grids & Epoch 2 valid grids & Common grids \\
\midrule
0.1 m & 82,059 & 79,162 & 77,719 \\
\bottomrule
\end{tabular*}
\end{table}

\begin{figure}[pos=H]
  \centering
  \includegraphics[width=\figwidth,height=.68\textheight,keepaspectratio]{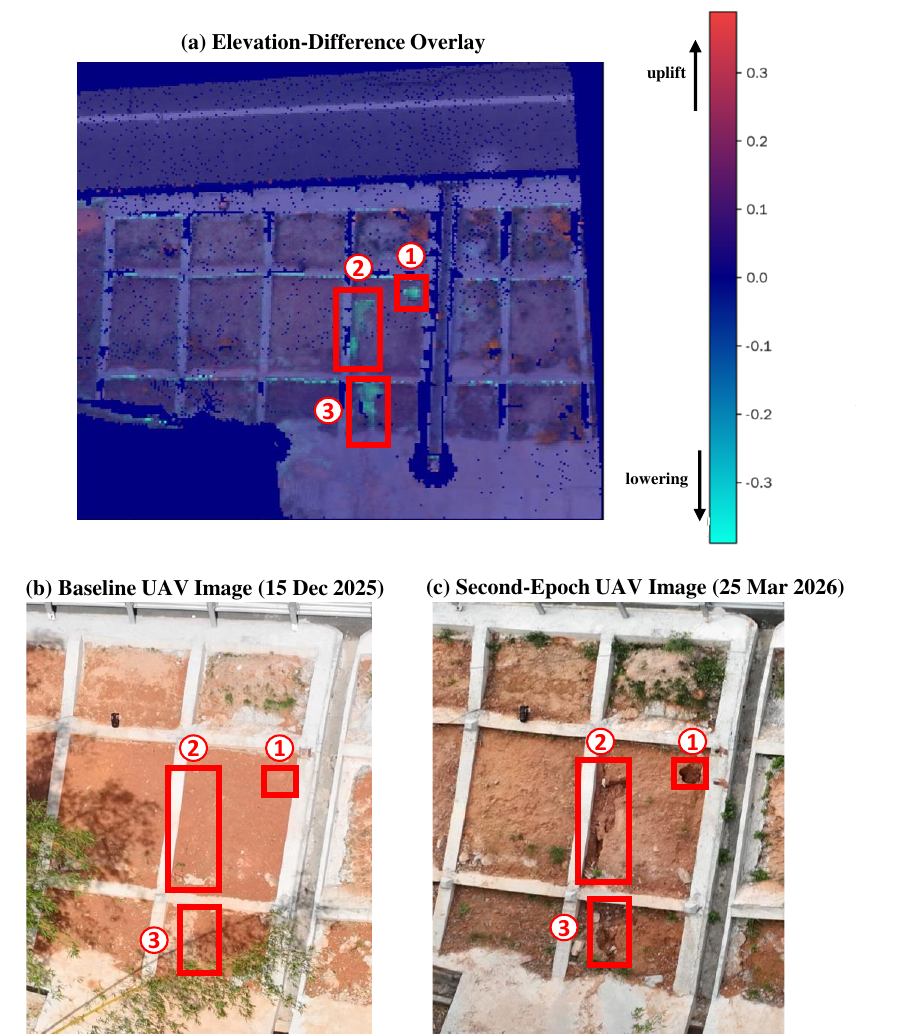}
  \caption{Elevation-difference overlay and time-separated UAV images of the real expressway slope case, showing spatial correspondence between negative elevation-change regions and visible erosion or material loss.}\label{fig:14}
\end{figure}

The resulting elevation-difference overlay is shown in Fig.~\ref{fig:14}. The heatmap detected a concentrated and contiguous negative elevation-change region in the exposed soil area between lattice beams, with values reaching approximately $-0.3~\mathrm{m}$. Because the response forms a spatially coherent patch and corresponds to visible erosion and material loss in the UAV imagery, it is more consistent with real surface degradation than with isolated differencing noise.

\subsection{Summary of Experimental Evidence}

The experiments provide three levels of evidence. First, the under-canopy foam-board experiment confirms that the shared preprocessing stage can extract usable ground-surface information under vegetation cover. Second, the simulation-based single-observation experiment shows that local uplift and subsidence regions can be preliminarily localized from one ground-surface point cloud, while also revealing drainage-channel false positives that require more field-specific training samples. Third, the controlled and real-field multi-epoch experiments show that grid-wise elevation differencing can quantify known artificial changes and localize real erosion-related surface loss. Together, these results support the feasibility of the proposed UAV-borne LiDAR workflow for automated slope-hazard monitoring.

\section{Discussion and Conclusion}\label{sec:discussion-conclusion}

\subsection{Discussion}

\subsubsection{Effectiveness of Ground-Surface Extraction}

The under-canopy foam-board experiment demonstrates the effectiveness of the ground-surface extraction stage in the proposed workflow. Although tree-canopy occlusion strongly interfered with direct observation of the ground surface, UAV-borne LiDAR still acquired valid returns from the foam-board target and its surrounding area. After CSF processing, a large proportion of canopy and other above-ground returns was suppressed, and the low-height foam-board target became more distinguishable in the filtered point cloud.

The elevation histogram further supports this observation. After CSF filtering, the elevation distribution showed reduced influence from high-elevation returns, indicating that non-ground points were effectively suppressed. These results show that the shared preprocessing stage can provide usable ground-surface information under canopy occlusion. At the same time, the extracted ground points were not spatially complete, which indicates that dense vegetation may still cause local holes and discontinuities. Therefore, this stage should be interpreted as a practical ground-surface extraction procedure rather than a complete reconstruction of all obscured terrain details.

\subsubsection{Feasibility of Single-Observation Hazard Screening}

The simulation experiment validates the preliminary feasibility of the single-observation hazard screening branch. Under the scene-level 5-fold cross-validation setting, the RandLA-Net-based branch achieved a mean OA of 64.71\% and a mean mIoU of 46.48\%. These results indicate that the model can learn useful geometric patterns for distinguishing normal background, uplift hazards, and depression hazards from single-epoch slope point clouds.

The slope rotation-flattening preprocessing is an important part of this branch because it reduces the interference of the global inclined slope background. By approximately aligning the principal slope plane with the horizontal plane, local uplift and depression anomalies become more distinguishable in the network input. This supports the use of geometric preprocessing before point-wise semantic recognition.

However, the current result should be regarded as a feasibility validation rather than a complete engineering-level verification. The dataset used in this experiment was generated through simulation, and real slope hazards may exhibit more complex shapes, unclear boundaries, and stronger background interference. Therefore, the single-observation branch provides a preliminary no-baseline screening capability, but further validation with real annotated slope-hazard point clouds is still needed.

\subsubsection{Reliability of Multi-Epoch Deformation Monitoring}

The multi-epoch deformation monitoring branch was validated through both controlled and real-field experiments. In the controlled known-deformation experiment, the gridded differencing method successfully detected the known artificial elevation changes. The square flat-topped uplift with a measured height of approximately 16 cm was represented by a mean elevation difference of 13.31 cm across 100 included grid cells. The circular conical uplift with a measured peak height of approximately 36 cm produced a maximum grid-wise elevation difference of 33.9 cm across 87 included grid cells. These results demonstrate that the gridded elevation differencing method has quantitative reliability under controlled conditions.

In the real expressway slope experiment, the two time-separated UAV-borne LiDAR surveys achieved a common grid coverage rate of 94.71\%, indicating good spatial consistency between the two epochs after preprocessing. The elevation-difference heatmap detected a concentrated negative elevation-change region with values reaching approximately $-0.3~\mathrm{m}$. This region was spatially consistent with the erosion and material-loss area observed in the UAV visible-light imagery. Therefore, the multi-epoch branch can provide spatially interpretable evidence for locating progressive deformation regions in real expressway slope environments.

\subsubsection{Complementarity of the Two Analytical Branches}

The two analytical branches play different roles in the proposed workflow. The single-observation branch is suitable for first inspection, emergency investigation, or no-baseline scenarios, where only one UAV-borne LiDAR survey is available. In contrast, the multi-epoch branch is suitable when time-separated observations have been accumulated and quantitative deformation analysis becomes possible.

These two branches are therefore complementary rather than redundant. The single-observation branch provides immediate hazard screening from one ground-surface point cloud, while the multi-epoch branch provides quantitative temporal evidence by comparing different survey epochs. Because both branches use the ground-surface point cloud extracted by the shared preprocessing stage, they can be integrated into a unified UAV-borne LiDAR-based slope-inspection workflow.

\subsubsection{Engineering Applicability of the Proposed Workflow}

The proposed workflow converts raw UAV-borne LiDAR data into two types of hazard-related outputs: a point-wise semantic segmentation map and an elevation-difference heatmap. These outputs are spatially interpretable and can help inspection personnel locate potential hazard areas more efficiently.

Compared with isolated processing components, the workflow better matches practical expressway slope inspection requirements. It supports both first-time hazard screening and multi-epoch deformation monitoring, and it can be implemented using standard UAV-borne LiDAR acquisition, point-cloud preprocessing, semantic segmentation, and grid-based differencing procedures. Therefore, the workflow provides an implementable technical route for automated slope-hazard monitoring and intelligent early warning.

\subsubsection{Limitations and Future Work}

Several limitations remain. First, the ground-surface extraction stage depends on the availability of valid ground returns. Under extremely dense vegetation, the number of laser pulses reaching the ground may still be insufficient, leading to holes and discontinuities in the extracted ground-surface point cloud. In addition, CSF parameters may need to be adjusted for different terrain, vegetation, and slope conditions, and local misclassification may occur in complex scenes.

Second, the single-observation hazard screening branch is currently validated using simulated slope-hazard data. Real hazard morphology may be more irregular, and point-wise annotations of real slope hazards are still limited. Future work should collect more real annotated UAV-borne LiDAR slope datasets and conduct more comprehensive comparisons with other point-cloud segmentation models.

Third, the multi-epoch deformation monitoring branch depends on registration quality and grid-size selection. Seasonal vegetation variation, soil moisture changes, root disturbance, and local gravel movement may introduce pseudo-deformation signals. The grid size (s) also affects the balance between sensitivity and smoothness. Future work may incorporate uncertainty estimation, adaptive grid strategies, or direct point-cloud comparison methods to improve the robustness of deformation analysis.

\subsection{Conclusion}

This study proposed an end-to-end UAV-borne LiDAR-based workflow for expressway slope inspection. The workflow first establishes a shared ground-surface representation from raw UAV-borne LiDAR point clouds and then supports two complementary analytical branches: single-observation hazard screening and multi-epoch deformation monitoring.

The ground-surface extraction stage was validated using the under-canopy foam-board experiment. The results show that UAV-borne LiDAR combined with CSF filtering can suppress a large proportion of above-ground returns and retain usable ground-surface information under canopy occlusion. The elevation-distribution statistics further support the effectiveness of CSF filtering in reducing the influence of non-ground points.

For single-observation hazard screening, the RandLA-Net-based branch was evaluated using simulated slope-hazard scenes. The model achieved a mean OA of 64.71\% and a mean mIoU of 46.48\% under scene-level 5-fold cross-validation, providing a preliminary baseline for no-baseline point-cloud hazard screening. For multi-epoch deformation monitoring, the gridded differencing method detected the known artificial uplift regions in the controlled experiment and identified a $-0.3~\mathrm{m}$ negative elevation-change region in a real expressway slope case, which corresponded spatially to visible erosion and material loss.

Overall, the proposed workflow provides an implementable solution for transforming UAV-borne LiDAR point clouds into interpretable slope-hazard outputs, including semantic segmentation maps and elevation-difference heatmaps. Future work will focus on collecting real annotated slope-hazard point clouds, improving robustness under dense vegetation, and integrating uncertainty-aware deformation analysis.

\bibliographystyle{cas-model2-names}
\bibliography{references}

\end{document}